%% file: main.tex
\documentclass[letterpaper]{article}

\input{packages}
\input{environment}

\title{Parallel bootstrap-based on-policy deep reinforcement learning for continuous fluid flow control applications}

\def\size{7.3cm}
\author{
	\parbox{\size}{\centering J. Viquerat\thanks{Corresponding author}}\\
	MINES Paristech, CEMEF\\
	PSL - Research University\\
	\texttt{jonathan.viquerat@mines-paristech.fr}\\
\And
	\parbox{\size}{\centering E. Hachem}\\
	MINES Paristech, CEMEF\\
	PSL - Research University
}

\begin{document}
\newgeometry{left=3cm,right=3cm,top=3cm,bottom=2.5cm}
\maketitle

\begin{abstract}
The coupling of deep reinforcement learning to numerical flow control problems has recently received a considerable attention, leading to groundbreaking results and opening new perspectives for the domain. Due to the usually high computational cost of fluid dynamics solvers, the use of parallel environments during the learning process represents an essential ingredient to attain efficient control in a reasonable time. Yet, most of the deep reinforcement learning literature for flow control relies on on-policy algorithms, for which the massively parallel transition collection may break theoretical assumptions and lead to suboptimal control models. To overcome this issue, we propose a parallelism pattern relying on partial-trajectory buffers terminated by a return bootstrapping step, allowing a flexible use of parallel environments while preserving the on-policiness of the updates. This approach is illustrated on a CPU-intensive continuous flow control problem from the literature.
\end{abstract}


\section{Introduction}

Deep neural networks (DNNs) have become a pervasive approach in a large variety of scientific domains in the course of the last decade, achieving multiple breakthroughs in domains such as image classification tasks \cite{rawat2017, khan2020}, speech recognition \cite{nassif2019} or generative tasks \cite{gui2020, ramesh2022}. Thanks to cheaper hardware and generalized access to large computational resources, such advances have led to a general evolution of the reference methods at both academic and industrial levels.

Among these progresses, decision-making techniques have largely benefited form the coupling of DNNs with reinforcement learning algorithms (called deep reinforcement learning, or DRL), due to their feature extraction capabilities and their ability to handle high-dimensional state spaces. Unprecedented efficiency has been achieved in many domains such as robotics \cite{pinto2017}, language processing \cite{bahdanau2016}, or games \cite{mnih2013, silver2017}, but also in the context of industrial applications \cite{kendall2018, bewley2018, googleDataCenter2018}.

In the recent years, DRL-based approaches have made their way into the domain of flow control, with an increasing amount of contributions on varied topics, such as (but not limited to) drag reduction \cite{rabault2019}, collective swimming \cite{novati2017} or heat transfers \cite{beintema2020}. Although a handful of couplings of DRL with experimental setups were reported, most contributions make use of numerical environments relying on a computational fluid dynamics (CFD) solvers \cite{garnier2021, viquerat2021}, making the performance of the latter an important lever for the successful learning of a control strategy by a DRL agent. Yet, in order to control processes of increasing computational loads, the use of parallel environments to accelerate the sample collection between agent updates also appears as a key ingredient, allowing to further exploit existing resources and reducing the time-to-control (here defined as the computational time required to reach efficient control). In 2019, such setup was introduced by Rabault and Kuhnle in the context of flow control for the drag reduction on a cylinder at low Reynolds number \cite{rabaultkuhnle2019}. In this contribution, the authors showed that, for a configuration where the CFD time represents almost 100\% of the time-to-control, using parallel collection of samples could yield an excellent speedup, up to the point where the number of parallel environments (hereafter denoted as $n_\text{env}$) becomes equal to the number of full episodes used for a single update (hereafter denoted as $n_\text{update}$, in this case equal to 20 episodes). Pushing the parallelization process beyond this point resulted in "over-parallelization" (\textit{sic}) and led to observables flat steps in the learning process (see \cite{rabaultkuhnle2019}, figure 4).

Although efficient, the approach introduced above comprises two major drawbacks. First, the "over-parallelization process" inherently leads to off-policy updates of the agent \blue{(\textit{i.e.} the agent is trained with samples produced by a previous policy)}, which introduces bias in the policy gradient estimates and increases the risk of the agent falling into local optima. While importance sampling corrections could be used to account for the discrepancy between the current policy and the behavior policy, such approach is known to introduce large variance in the estimate and require additional tuning \cite{meteli2018, tomczak2019}. Second, in order to take maximum advantage of the parallel sample collection, this approach constrains the size of the buffer update to be (i) a multiple of the number of transitions within an episode, and (ii) a multiple of the number of parallel environments. These two points may lead either to a suboptimal control model or to a suboptimal use of the available computational resources. To overcome these issues, we propose an update approach based on partial trajectories terminated by a return bootstrapping step to mimic a continuing environment \blue{(\textit{i.e.} the last reward of each parallel update buffer is modified to account for the continuous nature of the control task)}. The concept of partial trajectories is not new and was, for example, presented by Schulman in the original \blue{proximal policy optimization (PPO)} paper (\cite{ppo}, section 5), while the bootstrapping method was originally proposed by Pardo \textit{et al.} \cite{pardo2017}. As will be shown below, the combination of these two approaches allows to design a flexible parallel sample collection pattern, while retaining the on-policiness of the PPO algorithm \blue{(\textit{i.e.} the agent is trained with samples produced by the current policy)}. This represents a major asset in the context of fluid flow control, where efficient parallel sample collection eventually allows the control of complex, CPU-intensive environments.

The present paper is organized as follows: in section \ref{section:on-policy}, a short reminder of the basics of on-policy DRL algorithms and proximal policy optimization is provided. Then, a recall is made on the bootstrapping technique in section \ref{section:bootstrapping}, after what the proposed bootstrapped partial trajectories approach is detailed in section \ref{section:parallel}. To benchmark the proposed method, a continuous flow control case from the literature is presented in section \ref{section:cases} (namely the control of a falling fluid film, adapted from \cite{belus2019}). This case is then exploited in section \ref{section:results} to benchmark (i) the interest of bootstrapping and (ii) the proposed parallel paradigm. Finally, conclusions and perspectives are given.

\section{Parallel bootstrap-based on-policy deep reinforcement learning}

\subsection{On-policy DRL algorithms}
\label{section:on-policy}

Policy-based methods maximize the expected discounted cumulative reward of a policy $\pi(a \vert s)$ mapping states to actions, resorting to a probability distribution over actions given states. Among these techniques, two types of algorithms must be differentiated: 

\begin{enumerate}
	\item \emph{on-policy} algorithms usually require that the samples they use for training are generated with the current policy. In other words, after having collected a batch of transitions using the current policy, this data is used to update the agent and cannot be re-used for future updates;
	\item \emph{off-policy} algorithms are able to train on samples that were not collected with their current policy. They usually use a replay buffer to store transitions, and randomly sample mini-batches from it to perform updates. Hence, samples collected with a given policy can be re-used multiple times during the training procedure.
	
\end{enumerate}

Although off-policy algorithms naturally exhibit a better sample efficiency, their stability is not guaranteed, and, as of today, they have not yet made their way in the DRL-based flow control domain \cite{viquerat2021}. Contrarily, although less sample-efficient, on-policy algorithms present good stability properties, ease of implementation and tuning, and are widely represented in the field of DRL-based flow control. By focusing on on-policy algorithms, the present study can deliver more useful insights to the community. In the following section, a focus is made on the details of the well-known PPO algorithm \cite{ppo}.


\subsection{Proximal policy optimization (PPO)}
\label{section:ppo}

The PPO algorithm belongs to the class of actor-critic methods \cite{a2c}, in which an actor provides actions based on observations by sampling from a parameterized policy $\pi_\theta$, while a critic is devoted to evaluate the future expected discounted cumulative reward. The standard loss used to update the actor network reads:

\begin{equation*}
	L(\theta) = \underset{\tau \sim \pi_\theta}{\mathbb{E}} \left[ \sum_{t=0}^T \log \left( \pi_\theta (a_t \vert s_t) \right) A^{\pi_\theta} (s_t, a_t) \right],
\end{equation*}

\noindent where $A^{\pi_\theta}(s,a)$ is the advantage function, that represents the improvement in the expected cumulative reward when taking action $a$ in state $s$, compared to the average of all possible actions taken in state $s$. The advantage function is evaluated thanks to the rewards collected from the environment, and the evaluation of the value function provided by the critic. Although displaying good performance, vanilla actor-critic techniques displayed a high sensitivity to the learning rate, with small learning rates implying slow learning, while large learning rates leading to possible performance collapses. To overcome these issues, the proximal policy optimization \cite{ppo} uses a simple yet effective heuristic that helps avoid destructive updates. Namely, it relies on a clipped surrogate loss:

\begin{equation*}
	L(\theta) = \underset{(s,a) \sim \pi_{\theta_\text{old}}}{\mathbb{E}} \left[ \min \left( \frac{\pi_\theta (a \vert s)}{\pi_{\theta_{old}} (a \vert s)} , g \left( \epsilon, A^{\pi_{\theta_\text{old}}} (s,a) \right) \right) A^{\pi_{\theta_\text{old}}} (s,a) \right],
\end{equation*}

where

\begin{equation*}
	g(\epsilon, A) = 
	\begin{cases}
		(1+\epsilon) A 	& \text{if } A \geq 0,\\
		(1-\epsilon) A 	& \text{if } A < 0,
	\end{cases}
\end{equation*}

and $\epsilon$ is the clipping range, a small user-defined parameter defining how far away the new policy is allowed to go from the old one. Due to its improved learning stability and its relatively robust behaviour with respect to hyper-parameters, the PPO algorithm has received considerable attention in the DRL community, including in the context of flow control \cite{viquerat2021}.

It seems worth noting that, in practice, although PPO is considered an on-policy method in the sense that it does not make use of a replay buffer, it is not \emph{strictly} on-policy, as the collected transitions will be used for multiple epochs \emph{in the course of a single update} before being discarded. This fact leads a part of the community to label PPO as an off-policy method, which is fundamentally correct, although in essence the method is fundamentally different from state-of-the-art off-policy methods such as deep deterministic policy gradients-like methods (DDPG \cite{ddpg}, TD3 \cite{td3}).

\subsection{Bootstrapping}
\label{section:bootstrapping}

DRL methods can be used either to tackle \emph{episodic} or \emph{continuous} tasks. In episodic tasks, the agent has a limited amount of time to successfully reach a given goal, the provided time limit being an intrinsic characteristic of the problem (for example, the case of a robotic arm moving an object from an initial position to a final position). In continuous tasks, the time limit is often arbitrary, the control being supposed to run indefinitely (for example, the case of active drag reduction around an obstacle). In the case of continuous control, the time limit is set only so the environment can be regularly reset during the training, leading to an improved diversity of samples and a good representation of all the stages of control. In the context of on-policy algorithms such as PPO, this distinction leads to a subtle yet capital difference in the way the advantage buffer must be terminated when an episode is ended for a time-out reason \cite{pardo2017}. Traditionally, once an episode is terminated, the advantage vector is assembled using the reward vector $\bm{r}$ collected from the environment during the episode, the estimated value vector $\bm{\hat{v}_{\pi_\theta}}$ that estimates the discounted cumulative reward starting from a given state until the end of the episode, and the termination mask $\bm{m}$, which values are equal to $1$ for each non-terminal state, and $0$ for terminal states (so on a single episode, the termination mask is typically a vector of $1$ terminated with a $0$). Then, the advantage vector can be assembled as shown in algorithm \ref{alg:adv}.

\begin{algorithm}
\caption{Traditional advantage vector assembly}
\label{alg:adv}
\begin{algorithmic}[1]
\State \textbf{given}: the reward vector $\bm{r}$
\State \textbf{given}: the value vector $\bm{\hat{v}_{\pi_\theta}}$
\State \textbf{given}: the termination mask $\bm{m}$
\For{$t=0, n_{\text{steps}} -1$}
	\IIf{$y_t = r_t$}{$s_t$ is terminal}{$y_t = r_t + \gamma \hat{v}_{\pi_\theta} (s_{t+1})$}
\EndFor
\For{$t= n_{\text{steps}} -2, 0$}
	\State $y_t = y_t + \gamma  m_t y_{t+1}$
\EndFor
\State $\bm{A} = \bm{y} - \bm{\hat{v}_{\pi_\theta}}$
\end{algorithmic}
\end{algorithm} 

The computed advantage vector is later used in the computation of the actor loss\footnote{Here, the computation of the standard advantage function is presented for simplicity, yet the comments made in this section still hold for more complex advantage functions, such as the generalized advantage estimate \cite{gae}, for example.}. Yet, the algorithm \ref{alg:adv} does not account for terminations due to time limits in the context of continuous tasks. Hence, when trained, providing such an advantage vector to the agent does not account for the possible future rewards that could have been experienced if a different arbitrary time limit had been used. More, this approach inherently brings a credit attribution problem: reaching a similar state in the course of an episode or on a time-out termination would lead to very different outcomes that would be evaluated with the same inaccurate value function estimate $\hat{v}_{\pi_\theta} (s_t)$, thus leading to a degraded learning process. However, a fairly simply remedy, called \emph{bootstrapping}, consists in replacing the "\textbf{if} $s_t$ is terminal" condition of algorithm \ref{alg:adv} by "\textbf{if} $s_t$ is terminal \textbf{and} not a time-out". Indeed, one can clearly see from algorithm \ref{alg:adv} at line $8$ that modifying the final target value $y_t$ in the case of a time-out (\textit{i.e.} $t=T$) will modify all the previous target values $y_{t<T}$, thus having a large impact on the actor update, and eventually leading to a significant performance improvement. Experiments using bootstrapping alone in the context of continuous flow control tasks are presented in section \ref{section:bootstrapping_flow_results}. Additional results on regular benchmark control tasks from the \textsc{gym} and \textsc{mujoco} packages are also provided in appendix \ref{appendix:bootstrapping}.

\subsection{Parallel bootstrap-based learning}
\label{section:parallel}

In the context of CPU-expensive CFD environments, speeding up the training of DRL agents by harnessing the capabilities of parallelism can lead to substantial gains in computational resources. Although this can be done by exploiting the inner parallelism of the CFD computation itself, collecting data from environments running in parallel is also a well-known technique that has largely spread in the community. In the context of the coupling of CFD environments with DRL, such improvement represents a key ingredient for the discovery of efficient control laws, as the environment can represent from 80\% to more than 99\% of the computational time. In \cite{rabaultkuhnle2019}, the authors consider a 2D drag reduction case using the PPO algorithm, setting the update frequency of their agent to $n_\text{update} = 20$ episodes. Using an asynchronous parallelism of the environments, the authors report an excellent speedup up to $n_\text{env} = 20$ parallel environments, and a decent performance improvement up to 60. Yet, for $n_\text{env} > n_\text{update}$, a fraction of the updates are naturally performed in an off-policy way, due to the fact that the samples used were produced using a different policy than that being updated. Although an appreciable speedup is still observed, this choice can lead to a degraded learning, as will be evidenced in section \ref{section:bootstrap_parallelism}. Here, we propose to exploit the bootstrapping technique presented above to design a synchronous parallel paradigm that respects the on-policy nature of PPO.

\begin{figure}
\centering
\begin{tikzpicture}[	node distance=1cm and 0cm, 
				dot/.style = {circle, fill=red, minimum size=5pt, inner sep=0pt, outer sep=0pt},]
				
	\def\b{blue4}
	\def\g{gray4}
	\def\bg{bluegray4}
	\def\o{orange4}
	\def\w{white}
	\def\gr{green4}
	\def\p{purple4}
	\def\t{teal4}
	\def\r{royal4}
	
	\def\cs{0.38}

	
	\matrix[mStyle, nodes={mWH={\cs}{\cs}}] (ep1_1) at (0,0) {
		\node[mL={\b}] {}; &
		\node[mN={\b}] {}; &
		\node[mN={\b}] {}; &
		\node[mR={\b}] {}; \\
	};
	
	\matrix[mStyle, nodes={mWH={\cs}{\cs}}, right=0.2cm of ep1_1] (ep2_1) {
		\node[mL={\o}] {}; &
		\node[mN={\o}] {}; &
		\node[mN={\o}] {}; &
		\node[mR={\o}] {}; \\
	};
	
	\node[left=0.1cm of ep1_1, align = right, text width=2cm, text centered] (lab) {1 env\\ \scriptsize (sequential)};
	
	\matrix[mStyle, nodes={mWH={\cs}{\cs}}, right=0.5cm of ep2_1] (r1) {
		\node[mL={\b}] {}; &
		\node[mN={\b}] {}; &
		\node[mN={\b}] {}; &
		\node[mN={\b}] {}; &
		\node[mN={\o}] {}; &
		\node[mN={\o}] {}; &
		\node[mN={\o}] {}; &
		\node[mR={\o}] {}; \\
	};
	
	\matrix[mStyle, nodes={mWH={\cs}{\cs}}, right=0.5cm of r1] (b1) {
		\node[mL={\b}] {}; &
		\node[mN={\b}] {}; &
		\node[mN={\b}] {}; &
		\node[mN={\b}] (bts1) {}; &
		\node[mN={\o}] {}; &
		\node[mN={\o}] {}; &
		\node[mN={\o}] {}; &
		\node[mR={\o}] (bts2) {}; \\
	};
	
	\node[dot] at (bts1.east) {};
	\node[dot] at (bts2.east) {};
	
	\node[right=0.1cm of b1, align = right] {\hphantom{1 env}};
	
	
	\matrix[mStyle, nodes={mWH={\cs}{\cs}}, below=0.5cm of ep1_1] (ep1_2) {
		\node[mL={\b}] {}; &
		\node[mN={\b}] {}; &
		\node[mN={\b}] {}; &
		\node[mR={\b}] {}; \\
	};
	
	\matrix[mStyle, nodes={mWH={\cs}{\cs}}, below=0cm of ep1_2] (ep2_2) {
		\node[mL={\o}] {}; &
		\node[mN={\o}] {}; &
		\node[mN={\o}] {}; &
		\node[mR={\o}] {}; \\
	};
	
	\node[] (mid2) at ($0.5*(ep1_2) + 0.5*(ep2_2)$) {};
	\node[align = right, text width=2cm, text centered] at ($(lab |- mid2)$) {2 envs\\ \scriptsize (parallel)};
	
	\matrix[mStyle, nodes={mWH={\cs}{\cs}}] (r2) at (r1 |- mid2) {
		\node[mL={\b}] {}; &
		\node[mN={\b}] {}; &
		\node[mN={\b}] {}; &
		\node[mN={\b}] {}; &
		\node[mN={\o}] {}; &
		\node[mN={\o}] {}; &
		\node[mN={\o}] {}; &
		\node[mR={\o}] {}; \\
	};
	
	\matrix[mStyle, nodes={mWH={\cs}{\cs}}] (b2) at (b1 |- mid2) {
		\node[mL={\b}] {}; &
		\node[mN={\b}] {}; &
		\node[mN={\b}] {}; &
		\node[mN={\b}] (bts1) {}; &
		\node[mN={\o}] {}; &
		\node[mN={\o}] {}; &
		\node[mN={\o}] {}; &
		\node[mR={\o}] (bts2) {}; \\
	};
	
	\node[dot] at (bts1.east) {};
	\node[dot] at (bts2.east) {};

	
	\matrix[mStyle, nodes={mWH={\cs}{\cs}}, below=0.5cm of ep2_2] (ep1_4) {
		\node[mL={\b}] {}; &
		\node[mN={\b}] {}; &
		\node[mN={\gr}] {}; &
		\node[mR={\gr}] {}; \\
	};
	
	\matrix[mStyle, nodes={mWH={\cs}{\cs}}, below=0cm of ep1_4] (ep2_4) {
		\node[mL={\o}] {}; &
		\node[mN={\o}] {}; &
		\node[mN={\p}] {}; &
		\node[mR={\p}] {}; \\
	};
	
	\matrix[mStyle, nodes={mWH={\cs}{\cs}}, below=0cm of ep2_4] (ep3_4) {
		\node[mL={\bg}] {}; &
		\node[mN={\bg}] {}; &
		\node[mN={\t}] {}; &
		\node[mR={\t}] {}; \\
	};
	
	\matrix[mStyle, nodes={mWH={\cs}{\cs}}, below=0cm of ep3_4] (ep4_4) {
		\node[mL={\g}] {}; &
		\node[mN={\g}] {}; &
		\node[mN={\r}] {}; &
		\node[mR={\r}] {}; \\
	};
	
	\node[] (mid4) at ($0.5*(ep1_4) + 0.5*(ep4_4)$) {};
	\node[] (mid41) at ($0.5*(ep1_4) + 0.5*(ep2_4)$) {};
	\node[] (mid42) at ($0.5*(ep3_4) + 0.5*(ep4_4)$) {};
	\node[align = right, text width=2cm, text centered] at ($(lab |- mid4)$) {4 envs\\ \scriptsize (parallel)};
	
	\matrix[mStyle, nodes={mWH={\cs}{\cs}}] (r4) at (r1 |- mid41) {
		\node[mL={\b}] {}; &
		\node[mN={\b}] {}; &
		\node[mN={\gr}] {}; &
		\node[mN={\gr}] {}; &
		\node[mN={\o}] {}; &
		\node[mN={\o}] {}; &
		\node[mN={\p}] {}; &
		\node[mR={\p}] {}; \\
	};
	
	\node[] (r4mid) at ($0.5*(r4.west) + 0.5*(r4.east)$) {};
	\node[yshift=0.1cm] at ($(r4mid |- r4.north)$) {\scriptsize On-policy update};
	
	\matrix[mStyle, nodes={mWH={\cs}{\cs}}] (r5) at (r1 |- mid42) {
		\node[mL={\bg}] {}; &
		\node[mN={\bg}] {}; &
		\node[mN={\t}] {}; &
		\node[mN={\t}] {}; &
		\node[mN={\g}] {}; &
		\node[mN={\g}] {}; &
		\node[mN={\r}] {}; &
		\node[mR={\r}] {}; \\
	};
	
	\node[] (r5mid) at ($0.5*(r5.west) + 0.5*(r5.east)$) {};
	\node[yshift=0.1cm, red] at ($(r5mid |- r5.north)$) {\scriptsize Off-policy update};
	
	\matrix[mStyle, nodes={mWH={\cs}{\cs}}] (b4) at (b1 |- mid41) {
		\node[mL={\b}] {}; &
		\node[mN={\b}] (bts1) {}; &
		\node[mN={\o}] {}; &
		\node[mN={\o}] (bts2) {}; &
		\node[mN={\bg}] {}; &
		\node[mN={\bg}] (bts3) {}; &
		\node[mN={\g}] {}; &
		\node[mR={\g}] (bts4) {}; \\
	};
	
	\node[] (b4mid) at ($0.5*(b4.west) + 0.5*(b4.east)$) {};
	\node[yshift=0.1cm] at ($(b4mid |- b4.north)$) {\scriptsize On-policy update};
	
	\matrix[mStyle, nodes={mWH={\cs}{\cs}}] (b5) at (b1 |- mid42) {
		\node[mL={\gr}] {}; &
		\node[mN={\gr}] (bts5) {}; &
		\node[mN={\p}] {}; &
		\node[mN={\p}] (bts6) {}; &
		\node[mN={\t}] {}; &
		\node[mN={\t}] (bts7) {}; &
		\node[mN={\r}] {}; &
		\node[mR={\r}] (bts8) {}; \\
	};
	
	\node[] (b5mid) at ($0.5*(b5.west) + 0.5*(b5.east)$) {};
	\node[yshift=0.1cm] at ($(b5mid |- b5.north)$) {\scriptsize On-policy update};
	
	\node[dot, black] at (bts1.east) {};
	\node[dot, black] at (bts2.east) {};
	\node[dot, black] at (bts3.east) {};
	\node[dot, black] at (bts4.east) {};
	\node[dot] at (bts5.east) {};
	\node[dot] at (bts6.east) {};
	\node[dot] at (bts7.east) {};
	\node[dot] at (bts8.east) {};
	
	\node[] (top) at ($(lab) + (0,0.75)$) {};
	\draw[thick] ($(ep1_1.west |- top)$) -- ($(ep2_1.east |- top)$) node[pos=0.5,anchor=south] {Sample collection};
	\draw[thick] ($(r1.west |- top)$) -- ($(r1.east |- top)$) node[pos=0.5,anchor=south] {Regular buffer};
	\draw[thick] ($(b1.west |- top)$) -- ($(b1.east |- top)$) node[pos=0.5,anchor=south] {Bootstrapped buffer};

\end{tikzpicture}
\caption{\textbf{Illustration of the use of fixed-length trajectory buffers with bootstrapping in the context of parallel sample collection,} assuming that an update requires the unrolling of two full episodes. The red dots indicate end-of-episode bootstrapping, while black dots indicate partial-trajectory bootstrapping.}
\label{fig:buffer_based}
\end{figure}
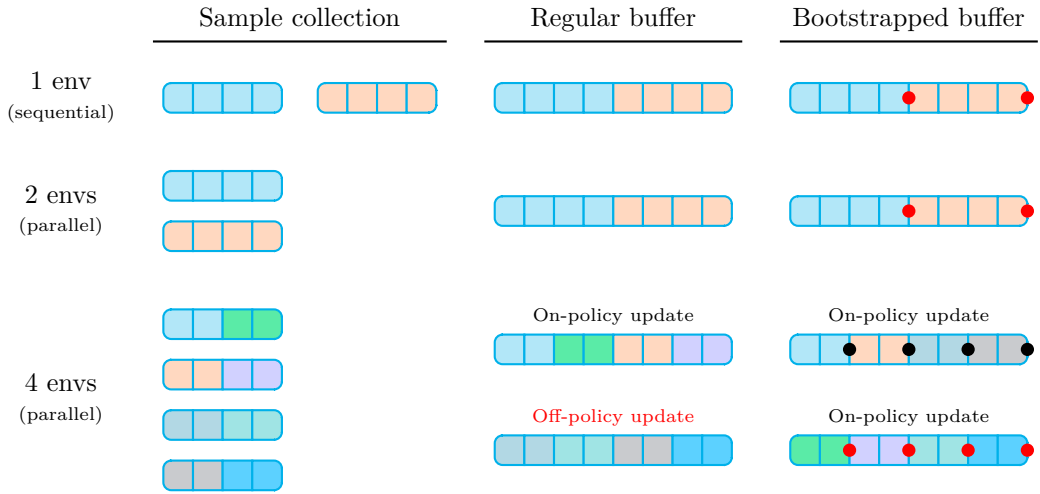

The proposed method is illustrated in figure \ref{fig:buffer_based} for a simple configuration: a regular episode consists of 4 transitions, and for this example it is decided that an update of the agent requires $n_\text{update} = 2$ full episodes, or 8 transitions. For $n_\text{env}=1$, two episodes are unrolled sequentially between each update. For $n_\text{env}=2$, transition sampling is effectively sped up while retaining the same update pattern, as two full episodes can be collected between each agent learning step. However, when reaching $n_\text{env}=4$, 4 full episodes are collected at once, and are then used to form two update buffers. While the first update will be on-policy, the second one will violate the on-policiness of the method, leading to possibly sub-optimal control. The proposed partial-trajectory approach holds two major differences: first, all collected transition buffers are terminated with a return bootstrap step; second, in the case of $n_\text{env} = 4$, only two transitions are unrolled for each environment, and a first buffer update is formed using 4 partial bootstrapped trajectories. Once the update is performed, the second half of the episodes is unrolled using the updated policy, forming a second update buffer used for a second on-policy update. For convenience, in the remaining of this paper, we call the first kind \emph{end-of-episode (EOE) bootstrapping}, while the second kind is designated as \emph{partial-trajectory (PT) bootstrapping}. Regarding the implementation of these features, adding EOE bootstrapping to an existing actor-critic code requires only minimal modifications, while PT bootstrapping requires a larger amount of modifications in the unrolling process.

In the context of real environments, this approach leads to an important flexibility, as (i) it does not require to collect full episodes to perform updates, (ii) it allows to better exploit parallel environments, and (iii) it preserves the on-policiness of the PPO algorithm. While it is important to notice that this method relies on a proper evaluation of the value function at the states at which bootstrapping is applied, our experiments show that the learning of the value function in the early steps of the process is sufficient for the method to bring an important benefit over the vanilla approach.

\section{Continuous flow control application}
\label{section:cases}

\begin{figure}
\centering
\def\env{\vphantom{t}env.}
\def\training{training}
\def\others{\vphantom{gt}others}
\begin{subfigure}[t]{.35\textwidth}
	\centering
	\begin{tikzpicture}[	trim axis left, trim axis right,font=\scriptsize]
	\begin{axis}[ 		ymin=0, ymax=105, ylabel={},
					ybar,
          				bar width=0.5cm,
         				bar shift=0pt, enlarge x limits={abs=0.5cm},
					symbolic x coords={\env, \training, \others},
					xtickmin=\env, xtickmax=\others, grid=both,
					width=\linewidth, height=4.5cm]

	\addplot[fill=orange1] 	coordinates {(\env, 22)};
	\addplot[fill=blue1] 		coordinates {(\training, 33) };
	\addplot[fill=gray1] 		coordinates {(\others, 45)};

	\end{axis}
	\end{tikzpicture}
	\caption{\textsc{pendulum-v1} environment}
	\label{fig:cpu_cost_pendulum}
\end{subfigure} \quad
\begin{subfigure}[t]{.35\textwidth}
	\centering
	\begin{tikzpicture}[	trim axis left, trim axis right,font=\scriptsize]
	\begin{axis}[ 		ymin=0, ymax=105, ylabel={},
					ybar,
          				bar width=0.5cm,
         				bar shift=0pt, enlarge x limits={abs=0.5cm},
					symbolic x coords={\env, \training, \others},
					xtickmin=\env, xtickmax=\others, grid=both,
					width=\linewidth, height=4.5cm]

	\addplot[fill=orange1] 	coordinates {(\env, 98)};
	\addplot[fill=blue1] 		coordinates {(\training, 0.004) };
	\addplot[fill=gray1] 		coordinates {(\others, 1.996)};

	\end{axis}
	\end{tikzpicture}
	\caption{Shkadov environment}
	\label{fig:cpu_cost_shkadov}
\end{subfigure} 
%
%
\caption{\textbf{Comparison of the CPU cost} between the simple \textsc{pendulum-v1} environment and the CPU-intensive task considered in this study. The "environment" part covers the stepping, the reward computation and the construction of the observations. The "training" part covers the data movements required for the construction of the training buffers, as well as the loss computation and the back-propagation. The "others" part covers all the remaining tasks, including mostly buffer management and handling of the parallelism.}
\label{fig:cpu_cost}
\end{figure}
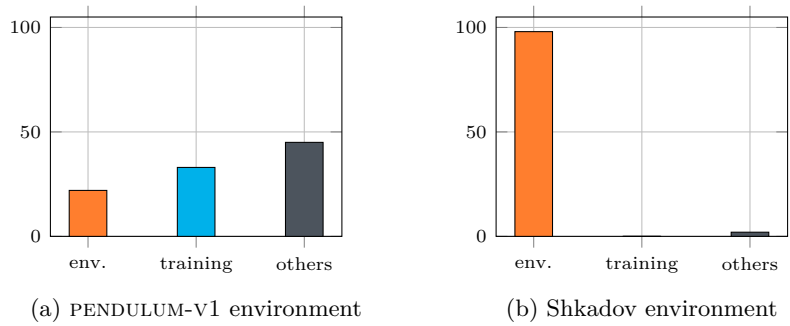

To illustrate the interest of the approach introduced in the previous section, a continuous flow control case from the literature is considered. To present meaningful results, the authors chose an environment for which the CPU cost of the unrolling is significantly higher than the training of the agent (typically around $96\%$ of the total CPU time). To illustrate, the repartition of the CPU time of these two cases is compared with that of the well-known \textsc{pendulum-v1} environment in figure \ref{fig:cpu_cost}. This section is devoted to the description of the cases, while the corresponding results are shown in section \ref{section:results}.

\subsection{Control of instabilities in a falling fluid film}

\begin{figure}
\centering
\pgfdeclarelayer{background}
\pgfsetlayers{background,main}
\begin{tikzpicture}[	scale=0.8, trim axis left, trim axis right, font=\scriptsize]
	\begin{axis}[	xmin=0, xmax=500, ymin=0.5, ymax=3.5, scale=1.0,
				width=\textwidth, height=.25\textwidth, scale only axis=true,
				legend cell align=left, legend pos=north east,
				grid=major, xlabel=$x$, ylabel=$h$]

		\begin{pgfonlayer}{background}
			\fill[color=bluegray3,opacity=0.3] (axis cs:0,0.5) rectangle (axis cs:150,5);
			\fill[color=orange3,opacity=0.3] (axis cs:150,0.5) rectangle (axis cs:275,5);
			\fill[color=teal3,opacity=0.3] (axis cs:275,0.5) rectangle (axis cs:500,5);
		\end{pgfonlayer}
		
		\addplot[draw=gray1, very thick, smooth] 			table[x index=0,y index=1] {fig/shkadov_free.dat};
		\addplot[draw=black, thick, dash pattern=on 2pt] 	coordinates {(150,0.5) (150,5)}; 
		\addplot[draw=black, thick, dash pattern=on 2pt] 	coordinates {(275,0.5) (275,5)};
			
	\end{axis}
\end{tikzpicture}
\caption{\textbf{Example of developed flow for the Shkadov equations with $\delta = 0.1$.} Three regions can be identified: a first region where the instability grows from a white noise (blue), a second region with pseudo-periodic waves (orange), and a third region with non-periodic, pulse-like waves (green).} 
\label{fig:shkadov_free}
\end{figure}
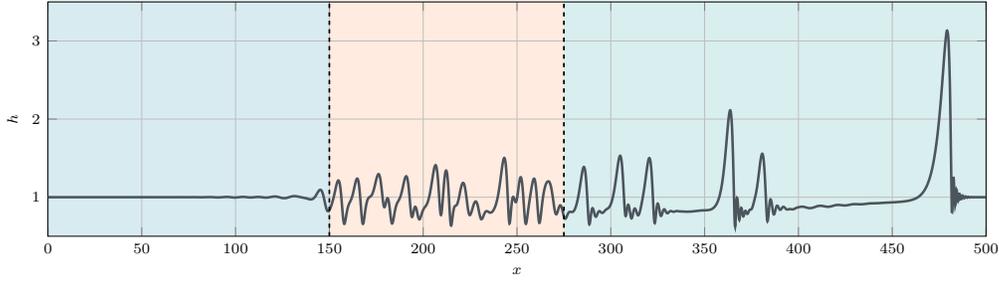

The considered case concerns the control of growing instabilities developing in a 1D falling fluid film perturbed with random noise, for which a simplified, two-equations model for this setup was proposed in 1967 by Shkadov \cite{shkadov1967}. Although it was found to lack some physical consistency \cite{lavalle2014}, this model displays interesting spatio-temporal dynamics while remaining acceptably cheap to integrate numerically. It simultaneously evolves the flow rate $q$ as well as the fluid height $h$ as:

\begin{equation}
\label{eq:shkadov}
\begin{split}
	\partial_t h 	&= -\partial_x q, \\
	\partial_t q		&= - \frac{6}{5} \partial_x \left( \frac{q^2}{h} \right) + \frac{1}{5 \delta} \left(  h \left( 1 + \partial_{xxx}h \right) - \frac{q}{h^2} \right),
\end{split}
\end{equation}

\noindent with all the physics of the problem being condensed in the $\delta$ parameter:

\begin{equation}
\label{eq:shkadov_delta}
	\delta = \frac{1}{15} \left( \frac{3 {Re}^2}{W} \right)^{\frac{1}{3}},
\end{equation}

\noindent where $Re$ and $W$ are the Reynolds and the Weber numbers, respectively defined on the flat-film thickness and the flat-film average velocity \cite{chang2002}. The system (\ref{eq:shkadov}) is solved on a 1D domain of length $L$, with the following initial and boundary conditions:

\begin{equation}
\label{eq:shkadov_bc}
\begin{split}
	q(x,0)	&= 1 \text{ and } h(x,0)	= 1, \\
	q(0,t) 	&= 1 \text{ and } h(0,t) 	= 1 + \mathcal{U} ( -\eps, \eps), \\
	\partial_x q(L,t) &= 0 \text{ and } 	\partial_x h(L,t) = 0,
\end{split}
\end{equation}

with $\eps \ll 1$ being the noise level. As shown in figure \ref{fig:shkadov_free}, the introduction of a random uniform noise at the inlet triggers the development of exponentially-growing instabilities (blue region) which eventually transition to a pseudo-periodic behavior (orange region). Then, the periodicity of the waves break, and the instabilities turn into into pulse-like structures, presenting a steep front preceded by small ripples \cite{chang2002book}. It is observed that some of these steep pulses, called solitary pulses, travel faster than others, and can capture upstream pulses in coalescence events. The dynamics of these solitary pulses are fully determined by the $\delta$ parameter, while the location of the transition regions also depends on the inlet noise level \cite{chang2002}.

The control environment proposed here is re-implemented based on the original publication of Belus \textit{et al.} \cite{belus2019}, although with some significant differences, noted hereafter. It is important to notice that the translational invariance feature introduced in \cite{belus2019} is \emph{not} exploited here. Equations (\ref{eq:shkadov}) are discretized using a finite difference approach. Due to the existence of sharp gradients, the convective terms are discretized using a TVD scheme with a minmod flux limiter. The discretized third-order derivative is obtained by chaining a second-order centered difference for the second derivative, chained with a second-order forward difference, leading a second-order approximation. Finally, the time derivatives are discretized using a second-order Adams-Bashforth method. \blue{The convergence of the numerical discretization is tested on a manufactured solution in appendix \ref{appendix:convergence}.}

\begin{figure}
\centering
\pgfdeclarelayer{background}
\pgfsetlayers{background,main}
\begin{tikzpicture}[	scale=0.8, trim axis left, trim axis right, font=\scriptsize]
	\begin{axis}[	xmin=0, xmax=217.5, ymin=0, ymax=2, scale=1.0,
				xtick={0,50,100,150,200,250,300},
				width=\textwidth, height=.25\textwidth, scale only axis=true,
				legend cell align=left, legend pos=north east,
				grid=major, xlabel=$x$, ylabel=$h$]
				
		\def\x{150}
		\def\w{1.5}
		\def\lobs{10}
		\def\lrwd{10}
		\def\s{10}
		
		\def\y{0.40}
		\def\h{0.02}
		\def\p{0.07}
		
		\draw[fill=green2, draw=gray2] 			(\x+0*\s-\w,1) rectangle (\x+0*\s+\w,1-0.089);
		\draw[fill=purple2, draw=gray2]			(\x+0*\s-\lobs,\y-0*\p-\h) rectangle (\x+0*\s,\y-0*\p+\h);
		\draw[fill=teal2, draw=gray2]			(\x+0*\s,\y-0*\p-\h) rectangle (\x+0*\s+\lrwd,\y-0*\p+\h);
		
		\draw[fill=green2, draw=gray2] 			(\x+1*\s-\w,1) rectangle (\x+1*\s+\w,1-0.343);
		\draw[fill=purple2, draw=gray2]			(\x+1*\s-\lobs,\y-1*\p-\h) rectangle (\x+1*\s,\y-1*\p+\h);
		\draw[fill=teal2, draw=gray2]			(\x+1*\s,\y-1*\p-\h) rectangle (\x+1*\s+\lrwd,\y-1*\p+\h);
		
		\draw[fill=green2, draw=gray2] 			(\x+2*\s-\w,1) rectangle (\x+2*\s+\w,1-0.602);
		\draw[fill=purple2, draw=gray2]			(\x+2*\s-\lobs,\y-2*\p-\h) rectangle (\x+2*\s,\y-2*\p+\h);
		\draw[fill=teal2, draw=gray2]			(\x+2*\s,\y-2*\p-\h) rectangle (\x+2*\s+\lrwd,\y-2*\p+\h);
		
		\draw[fill=green2, draw=gray2] 			(\x+3*\s-\w,1) rectangle (\x+3*\s+\w,1-0.403);
		\draw[fill=purple2, draw=gray2]			(\x+3*\s-\lobs,\y-3*\p-\h) rectangle (\x+3*\s,\y-3*\p+\h);
		\draw[fill=teal2, draw=gray2]			(\x+3*\s,\y-3*\p-\h) rectangle (\x+3*\s+\lrwd,\y-3*\p+\h);
		
		\draw[fill=green2, draw=gray2] 			(\x+4*\s-\w,1) rectangle (\x+4*\s+\w,1+0.284);
		\draw[fill=purple2, draw=gray2]			(\x+4*\s-\lobs,\y-4*\p-\h) rectangle (\x+4*\s,\y-4*\p+\h);
		\draw[fill=teal2, draw=gray2]			(\x+4*\s,\y-4*\p-\h) rectangle (\x+4*\s+\lrwd,\y-4*\p+\h);
		
		\addplot[draw=gray1, very thick, smooth] 	table[x index=0,y index=1] {fig/shkadov_control.dat};
			
	\end{axis}
\end{tikzpicture}
\caption{\textbf{Example of observation and reward computation areas} for 5 jets. The jets strengths are shown with green rectangles, while the observation areas (upstream of each jet) and reward areas (downstream of each jet) are shown in purple and teal, respectively.} 
\label{fig:shkadov_jets}
\end{figure}
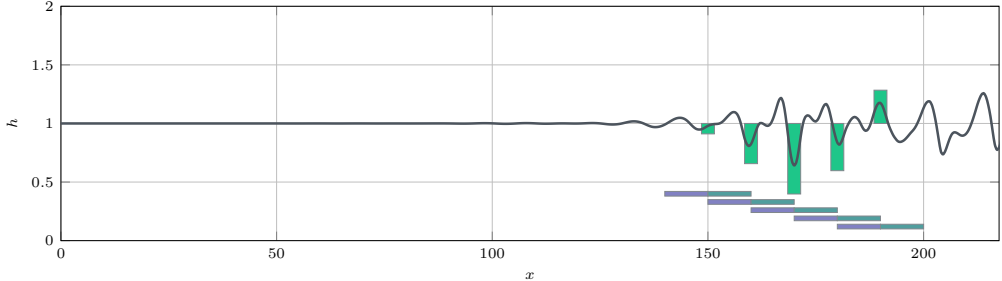

The control of the system (\ref{eq:shkadov}) is performed by adding a forcing term $\delta q_j$ to the equation driving the temporal evolution of the flow rate. In practice, this is achieved by adding localized jets at certain positions in the domain, as shown in figure \ref{fig:shkadov_jets}, which strengths are to be controlled by the DRL agent. The first jet is positioned by default at $x_0=150$, with jet spacing being by default set to $\Delta x_\text{jets} = 10$, similarly to \cite{belus2019}. To save computational time, the length of the domain is a function of the number of jets $n_\text{jets}$ and their spacing:

\begin{equation}
	L = L_0 + \left( n_\text{jets} + 2\right) \Delta x_\text{jets}.
\end{equation}

By default, $L_0 = 150$ (which corresponds to the start of the pseudo-periodic region for $\delta = 0.1$), and $n_\text{jets}$ is set equal to 1. The spatial discretization step is set as $\Delta x = 0.5$, while the numerical time step is $\Delta t = 0.005$. The inlet noise level is set as $\eps = \num{5e-4}$, similarly to \cite{belus2019}. The injected flow rate $\delta q_j$ has the following form:

\begin{equation}
\label{eq:shkadov_jets}
	\delta q_j (x,t) = A u_j(t) \frac{4 \left( x - x_j^l \right) \left( x_j^r - x \right)}{ \left( x_j^r - x_j^l \right)^2},
\end{equation}

with $A=5$ an \textit{ad-hoc} non-dimensional amplitude factor, $x_j^l$ and $x_j^r$ the left and right limits of jet $j$, and $u_j(t) \in [-1,1]$ the action provided by the agent. Expression (\ref{eq:shkadov_jets}) corresponds to a parabolic profile of the jet in $x$, such that the injected flow rate drops to $0$ on the boundaries of each jet. The jet width $x_j^r - x_j^l$ is set equal to $4$, similarly to \cite{belus2019}. The time dependance of $u(t)$ is implemented as a saturated linear variation from an action to the next one, in the form:

\begin{equation}
\label{eq:shkadov_actions}
	u_j(t) = (1-\alpha(t)) u_j^{n-1} + \alpha(t) u_j^n \text{, with } \alpha(t) = \min \left( \frac{t-t_n}{\Delta t_\text{int}}, 1 \right),
\end{equation}

Hence, when the actor provides a new action to the environment at time $t=t_n$, the real imposed action is a linear interpolation between the previous action $u_j^{n-1}$ and the new action $u_j^n$ over a time $\Delta t_\text{int}$ (here taken equal to $0.01$ time units), after what the new action is imposed over the remaining action time $\Delta t_\text{const}$ (here taken equal to $0.04$ time units). The total action time-step is therefore $\Delta t_\text{act} = \Delta t_\text{int} + \Delta t_\text{const}$, which value is therefore equal to $0.05$ time units. The total episode time is fixed to $20$ time units, corresponding to $400$ actions.

The observations provided to the agent are the \blue{mass flow rates} collected in the union of regions $A^j_\text{obs}$ of length $l_\text{obs} = 10$ located upstream of each jet, as shown in figure \ref{fig:shkadov_jets}. Contrarily to the original article, the flow rates of this region are not provided to the agent, and the observations are not clipped. The reward for each jet $j$ is computed on a region $A^j_\text{rwd}$ of length $l_\text{rwd} = 10$ located downstream of it (see figure \ref{fig:shkadov_jets}), the global reward consisting of a weighted sum of each individual reward:

\begin{equation}
\label{eq:shkadov_reward}
	r(t) = - \frac{1}{l_\text{rwd} \, n_\text{jets}} \sum_{j=0}^{n_\text{jets}-1} \sum_{x \in A^j_\text{rwd}} (h(x,t) - 1)^2
\end{equation}

Finally, each episode starts by randomly loading a fully developed initial state from a pre-computed set. The latter were obtained by solving the uncontrolled equations from an initial flat film configuration during a time $t_\text{init}$ comprise between $200$ and $220$ time units.

\subsection{Default PPO parameters}

The default parameters used for the present study are presented in table \ref{table:parameters}. A PPO agent is used with separate networks for the actor and the critic, the actions being drawn from a multivariate normal law with diagonal covariance matrix. While the critic network is a simple feedforward network with two hidden layers of size 64, the actor network is made of a trunk of size 64, with two branches composed of a single layer, each of size 64. The first branch is terminated using a tanh layer, used to output the mean of the normal distribution, while the second branch ends with a sigmoid layer, used to output the standard deviation of the distribution. The actions drawn from the corresponding distribution are clipped in $[-1,1]^d$ before being mapped to their adequate physical range. The generalized advantage estimate \cite{gae} is used, and the advantage vectors are normalized per rollout. Additionally, we underline that the current parallel implementation is based on the message-passing interface (MPI), which led to improved parallel speedups over shared-memory approach.

\begin{table}
    \footnotesize
    \caption{\textbf{Default parameters used in this study.}}
    \label{table:parameters}
    \centering
    \begin{tabular}{rll}
        \toprule
        --					& agent type					& PPO-clip\\
	$\gamma$ 			& discount factor				& 0.99\\
	$\lambda_a$ 			& actor learning rate				& \num{5e-4}\\
	$\lambda_c$ 			& critic learning rate				& \num{2e-3}\\
	--		 			& optimizer					& adam\\
	--					& weights initialization			& orthogonal\\
	--	 				& activation (hidden layers)		& relu\\
	-- 					& activation (actor final layer)		& tanh, sigmoid\\
	-- 					& activation (critic final layer)		& linear\\
	$\epsilon$ 			& PPO clip value				& 0.2\\
	$\beta$				& entropy bonus				& 0.01\\
	$g$					& gradient clipping value			& 0.1\\	
	-- 					& actor network					& $[64, [[64],[64]]]$\\
	-- 					& critic network					& $[64, 64]$\\
	--					& observation normalization		& yes\\
	--					& observation clipping			& no\\
	--					& advantage type				& GAE\\
	$\lambda_\text{GAE}$	& bias-variance trade-off			& 0.99\\
	--					& advantage normalization		& yes\\
        \bottomrule
    \end{tabular}
\end{table}

\section{Results}
\label{section:results}

In this section, we evaluate the interest of bootstrapping, as well as the performance of the proposed parallel paradigm against the canonical parallel approach. First, we simply compare the score curves obtained from sequential learning, with and without the end-of-episode (EOE) bootstrapping step. Then, we investigate the performance of the partial-trajectory (PT) bootstrapping parallel technique against the standard approach. Since they do not represent the core of this contribution, results from the solved environment are proposed in appendix \ref{appendix:solved_env}.

\subsection{End-of-episode bootstrapping}
\label{section:bootstrapping_flow_results}

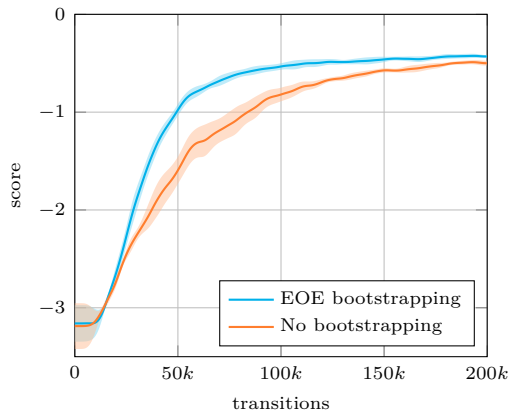
\begin{figure}
\centering
\def\nsteps{200000}
\begin{tikzpicture}[	trim axis left, trim axis right, font=\scriptsize,
				upper/.style={	name path=upper, smooth, draw=none},
				lower/.style={	name path=lower, smooth, draw=none},
				node distance=0.05cm]
	\begin{axis}[	xmin=0, xmax=200000, ymin=-3.5, ymax=0, scale=0.8,
				xtick={0,50000,100000,150000,200000},
				xticklabels={$0$,$50k$,$100k$,$150k$,$200k$},
				scaled x ticks=false,
				legend style={name=leg},
				every tick label/.append style={font=\scriptsize},
				legend cell align=left, legend pos=south east,
				grid=major, xlabel=transitions, ylabel={score}]
				
		\legend{EOE bootstrapping, No bootstrapping}
		
		\addplot [upper, forget plot] 				table[x index=0,y index=7] {fig/shkadov_bs.dat}; 
		\addplot [lower, forget plot] 				table[x index=0,y index=6] {fig/shkadov_bs.dat}; 
		\addplot [fill=blue3, opacity=0.5, forget plot] 	fill between[of=upper and lower];
		\addplot[draw=blue1, thick, smooth] 			table[x index=0,y index=5] {fig/shkadov_bs.dat};
		
		\addplot [upper, forget plot] 				table[x index=0,y index=7] {fig/shkadov_no_bs.dat}; 
		\addplot [lower, forget plot] 				table[x index=0,y index=6] {fig/shkadov_no_bs.dat}; 
		\addplot [fill=orange3, opacity=0.5, forget plot] 	fill between[of=upper and lower];
		\addplot[draw=orange1, thick, smooth] 		table[x index=0,y index=5] {fig/shkadov_no_bs.dat}; 
		
	\end{axis}
\end{tikzpicture}
\caption{\textbf{Comparison of score curves with and without end-of-episode bootstrapping} in a sequential learning context}
\label{fig:bootstrap_vs_regular}
\end{figure}

First, we consider the sole impact of EOE bootstrapping on the agent training. To do so, we compare the score curves obtained by training an agent in sequential mode on both environments, with "regular" ending (\textit{i.e.} no EOE bootstrapping step) and with bootstrapped ending. As can be observed in figure \ref{fig:bootstrap_vs_regular}, EOE bootstrapping accelerates the convergence of the agent while also reducing the variability in performance between the different runs. On the Shakdov environment, it almost cuts by half the number of required transitions to reach the maximal score. To illustrate further the benefits of EOE bootstrapping, results on standard \textsc{gym} environments are also presented in appendix \ref{appendix:bootstrapping}. Similar conclusions are drawn from these additional examples.

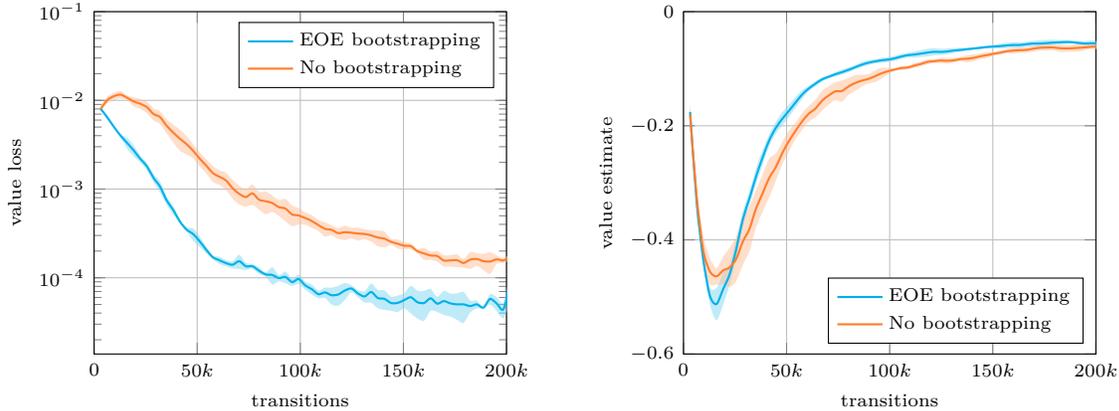
\begin{figure}
\centering
\def\nsteps{200000}
\begin{subfigure}[t]{.45\textwidth}
	\centering
	\begin{tikzpicture}[	trim axis left, trim axis right, font=\scriptsize,
					upper/.style={	name path=upper, smooth, draw=none},
					lower/.style={	name path=lower, smooth, draw=none},
					node distance=0.05cm]
		\begin{semilogyaxis}[	xmin=0, xmax=200000, ymin=0.0, ymax=0.1, scale=0.8,
					xtick={0,50000,100000,150000,200000},
					xticklabels={$0$,$50k$,$100k$,$150k$,$200k$},
					scaled x ticks=false,
					legend style={name=leg},
					every tick label/.append style={font=\scriptsize},
					legend cell align=left, legend pos=north east,
					grid=major, xlabel=transitions, ylabel={value loss}]
				
			\legend{EOE bootstrapping, No bootstrapping}
		
			\addplot [upper, forget plot] 				table[x index=0,y index=3] {fig/v_loss_bs.dat}; 
			\addplot [lower, forget plot] 				table[x index=0,y index=4] {fig/v_loss_bs.dat}; 
			\addplot [fill=blue3, opacity=0.5, forget plot] 	fill between[of=upper and lower];
			\addplot[draw=blue1, thick, smooth] 			table[x index=0,y index=2] {fig/v_loss_bs.dat};
		
			\addplot [upper, forget plot] 				table[x index=0,y index=3] {fig/v_loss_no_bs.dat}; 
			\addplot [lower, forget plot] 				table[x index=0,y index=4] {fig/v_loss_no_bs.dat}; 
			\addplot [fill=orange3, opacity=0.5, forget plot] 	fill between[of=upper and lower];	
			\addplot[draw=orange1, thick, smooth] 		table[x index=0,y index=2] {fig/v_loss_no_bs.dat}; 
		
		\end{semilogyaxis}
	\end{tikzpicture}
	\caption{Evolution of the value loss during training}
	\label{fig:value_loss}
\end{subfigure} \qquad
\begin{subfigure}[t]{.45\textwidth}
	\centering
	\begin{tikzpicture}[	trim axis left, trim axis right, font=\scriptsize,
					upper/.style={	name path=upper, smooth, draw=none},
					lower/.style={	name path=lower, smooth, draw=none},
					node distance=0.05cm]
		\begin{axis}[	xmin=0, xmax=200000, ymin=-0.6, ymax=0.0, scale=0.8,
					xtick={0,50000,100000,150000,200000},
					xticklabels={$0$,$50k$,$100k$,$150k$,$200k$},
					scaled x ticks=false,
					legend style={name=leg},
					every tick label/.append style={font=\scriptsize},
					legend cell align=left, legend pos=south east,
					grid=major, xlabel=transitions, ylabel={value estimate}]
				
			\legend{EOE bootstrapping, No bootstrapping}
		
			\addplot [upper, forget plot] 				table[x index=0,y index=6] {fig/v_loss_bs.dat}; 
			\addplot [lower, forget plot] 				table[x index=0,y index=7] {fig/v_loss_bs.dat}; 
			\addplot [fill=blue3, opacity=0.5, forget plot] 	fill between[of=upper and lower];
			\addplot[draw=blue1, thick, smooth] 			table[x index=0,y index=5] {fig/v_loss_bs.dat};
		
			\addplot [upper, forget plot] 				table[x index=0,y index=6] {fig/v_loss_no_bs.dat}; 
			\addplot [lower, forget plot] 				table[x index=0,y index=7] {fig/v_loss_no_bs.dat}; 
			\addplot [fill=orange3, opacity=0.5, forget plot] 	fill between[of=upper and lower];
			\addplot[draw=orange1, thick, smooth] 		table[x index=0,y index=5] {fig/v_loss_no_bs.dat}; 
		
		\end{axis}
	\end{tikzpicture}
	\caption{Evolution of the value estimate during training}
	\label{fig:value_estimate}
\end{subfigure}
\caption{\textbf{Evolution of the value loss and value estimate} during the course of training on the Shkadov environment, with and without end-of-episode bootstrapping.}
\label{fig:value_plots}
\end{figure}

To better understand the effects of the EOE bootstrapping, the plots of the value loss and value estimate along the course of training are proposed in figures \ref{fig:value_loss} and \ref{fig:value_estimate} respectively. A significantly lower value loss is observed, indicating that the EOE bootstrapping, by solving the credit assignment issue \cite{pardo2017}, induces a smoother value landscape for the critic to learn. This results in higher value estimates, which is expected due to the nature of the EOE bootstrapping procedure.

\subsection{Bootstrap-based parallelism}
\label{section:bootstrap_parallelism}

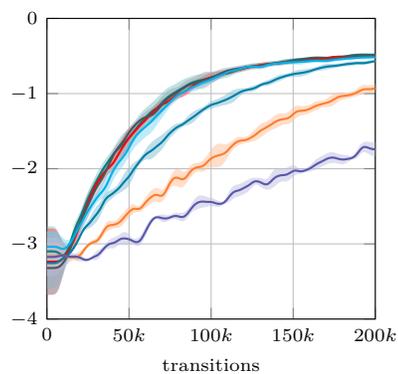
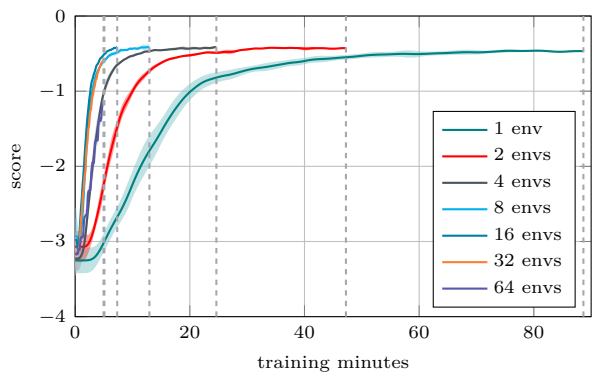
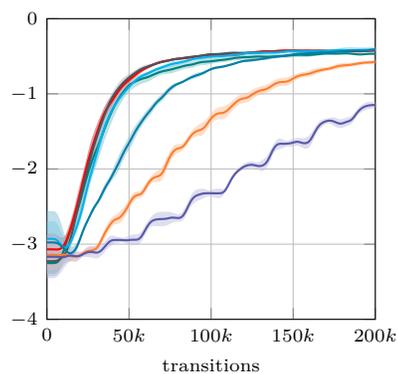
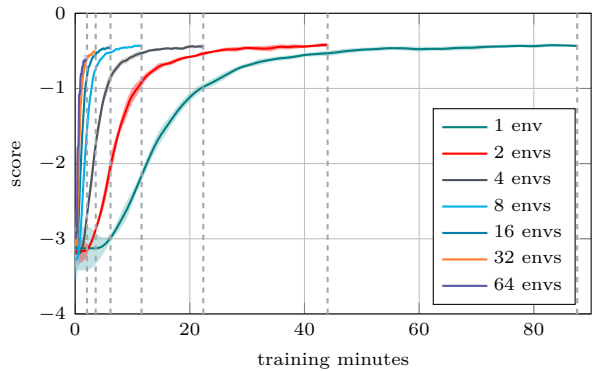
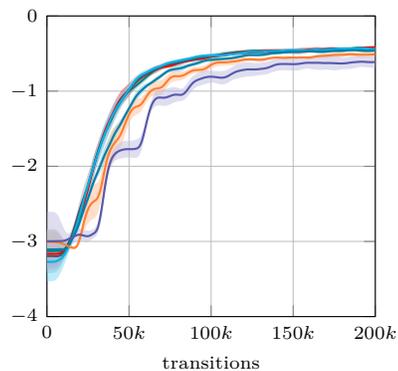
\begin{figure}[p]
\centering
\def\nsteps{200000}
\begin{subfigure}[t]{.55\textwidth}
\centering
	\begin{tikzpicture}[	trim axis left, trim axis right, font=\scriptsize,
					upper/.style={	name path=upper, smooth, draw=none},
					lower/.style={	name path=lower, smooth, draw=none},
					limit/.style={	gray3, thick, dash pattern=on 2pt}]
		\begin{axis}[	xmin=0, xmax=90, ymin=-4, ymax=0, scale=0.8,
					width=\textwidth, height=5cm, scale only axis=true,
					every tick label/.append style={font=\scriptsize},
					legend cell align=left, legend pos=south east,
					grid=major, xlabel=training minutes, ylabel=score]

			\legend{1 env, 2 envs, 4 envs, 8 envs, 16 envs, 32 envs, 64 envs}
	
			\def\tun{5380}
			\addplot [upper, forget plot] 				table[x expr=\thisrowno{0}/\nsteps*\tun/60,y index=7] {fig/shkadov_1_env_no_bs.dat}; 
			\addplot [lower, forget plot] 				table[x expr=\thisrowno{0}/\nsteps*\tun/60,y index=6] {fig/shkadov_1_env_no_bs.dat}; 
			\addplot [fill=teal3, opacity=0.5, forget plot] 	fill between[of=upper and lower];
			\addplot[draw=teal1, thick, smooth] 			table[x expr=\thisrowno{0}/\nsteps*\tun/60,y index=5] {fig/shkadov_1_env_no_bs.dat}; 
			\addplot[limit, forget plot]					coordinates {(\tun/60,-4) (\tun/60,0)};
		
			\def\ttwo{2834}
			\addplot [upper, forget plot] 				table[x expr=\thisrowno{0}/\nsteps*\ttwo/60,y index=7] {fig/shkadov_2_envs_no_bs.dat}; 
			\addplot [lower, forget plot] 				table[x expr=\thisrowno{0}/\nsteps*\ttwo/60,y index=6] {fig/shkadov_2_envs_no_bs.dat}; 
			\addplot [fill=red3, opacity=0.5, forget plot] 	fill between[of=upper and lower];
			\addplot[draw=red1, thick, smooth] 			table[x expr=\thisrowno{0}/\nsteps*\ttwo/60,y index=5] {fig/shkadov_2_envs_no_bs.dat};
			\addplot[limit, forget plot]					coordinates {(\ttwo/60,-4) (\ttwo/60,0)};
			
			\def\tfour{1470}
			\addplot [upper, forget plot] 				table[x expr=\thisrowno{0}/\nsteps*\tfour/60,y index=7] {fig/shkadov_4_envs_no_bs.dat}; 
			\addplot [lower, forget plot] 				table[x expr=\thisrowno{0}/\nsteps*\tfour/60,y index=6] {fig/shkadov_4_envs_no_bs.dat}; 
			\addplot [fill=gray3, opacity=0.5, forget plot] 	fill between[of=upper and lower];
			\addplot[draw=gray1, thick, smooth] 			table[x expr=\thisrowno{0}/\nsteps*\tfour/60,y index=5] {fig/shkadov_4_envs_no_bs.dat};
			\addplot[limit, forget plot]					coordinates {(\tfour/60,-4) (\tfour/60,0)};
		
			\def\thuit{775}
			\addplot [upper, forget plot] 				table[x expr=\thisrowno{0}/\nsteps*\thuit/60,y index=7] {fig/shkadov_8_envs_no_bs.dat}; 
			\addplot [lower, forget plot] 				table[x expr=\thisrowno{0}/\nsteps*\thuit/60,y index=6] {fig/shkadov_8_envs_no_bs.dat}; 
			\addplot [fill=blue3, opacity=0.5, forget plot] 	fill between[of=upper and lower];
			\addplot[draw=blue1, thick, smooth] 			table[x expr=\thisrowno{0}/\nsteps*\thuit/60,y index=5] {fig/shkadov_8_envs_no_bs.dat}; 
			\addplot[limit, forget plot]					coordinates {(\thuit/60,-4) (\thuit/60,0)};
		
			\def\tseize{441}
			\addplot [upper, forget plot] 					table[x expr=\thisrowno{0}/\nsteps*\tseize/60,y index=7] {fig/shkadov_16_envs_no_bs.dat}; 
			\addplot [lower, forget plot] 					table[x expr=\thisrowno{0}/\nsteps*\tseize/60,y index=6] {fig/shkadov_16_envs_no_bs.dat}; 
			\addplot [fill=bluegray3, opacity=0.5, forget plot] 	fill between[of=upper and lower];
			\addplot[draw=bluegray1, thick, smooth] 			table[x expr=\thisrowno{0}/\nsteps*\tseize/60,y index=5] {fig/shkadov_16_envs_no_bs.dat}; 
			\addplot[limit, forget plot]						coordinates {(\tseize/60,-4) (\tseize/60,0)};
			
			\def\ttdeux{304}
			\addplot [upper, forget plot] 				table[x expr=\thisrowno{0}/\nsteps*\ttdeux/60,y index=7] {fig/shkadov_32_envs_no_bs.dat}; 
			\addplot [lower, forget plot] 				table[x expr=\thisrowno{0}/\nsteps*\ttdeux/60,y index=6] {fig/shkadov_32_envs_no_bs.dat}; 
			\addplot [fill=orange3, opacity=0.5, forget plot] 	fill between[of=upper and lower];
			\addplot[draw=orange1, thick, smooth] 		table[x expr=\thisrowno{0}/\nsteps*\ttdeux/60,y index=5] {fig/shkadov_32_envs_no_bs.dat}; 
			\addplot[limit, forget plot]					coordinates {(\ttdeux/60,-4) (\ttdeux/60,0)};
			
			\def\tsquatre{293}
			\addplot [upper, forget plot] 				table[x expr=\thisrowno{0}/\nsteps*\tsquatre/60,y index=7] {fig/shkadov_64_envs_no_bs.dat}; 
			\addplot [lower, forget plot] 				table[x expr=\thisrowno{0}/\nsteps*\tsquatre/60,y index=6] {fig/shkadov_64_envs_no_bs.dat}; 
			\addplot [fill=purple3, opacity=0.5, forget plot] 	fill between[of=upper and lower];
			\addplot[draw=purple1, thick, smooth] 		table[x expr=\thisrowno{0}/\nsteps*\tsquatre/60,y index=5] {fig/shkadov_64_envs_no_bs.dat}; 
			\addplot[limit, forget plot]					coordinates {(\tsquatre/60,-4) (\tsquatre/60,0)};
			
		\end{axis}
	\end{tikzpicture}
	\caption{Score against walltime, no bootstrapping}
	\label{fig:shkadov_parallel_walltime_no_bootstrap}
\end{subfigure} \quad
\begin{subfigure}[t]{.35\textwidth}
\centering
	\begin{tikzpicture}[	trim axis left, trim axis right, font=\scriptsize,
					upper/.style={	name path=upper, smooth, draw=none},
					lower/.style={	name path=lower, smooth, draw=none},]
		\begin{axis}[	xmin=0, xmax=200000, ymin=-4, ymax=0, scale=0.8,
					xtick={0,50000,100000,150000,200000},
					xticklabels={$0$,$50k$,$100k$,$150k$,$200k$},
					scaled x ticks=false,
					width=\textwidth, height=5cm, scale only axis=true,
					every tick label/.append style={font=\scriptsize},
					legend cell align=left, legend pos=south east,
					grid=major, xlabel=transitions, ylabel={}]
				
			\addplot [upper, forget plot] 				table[x index=0,y index=7] {fig/shkadov_1_env_no_bs.dat};
			\addplot [lower, forget plot] 				table[x index=0,y index=6] {fig/shkadov_1_env_no_bs.dat}; 
			\addplot [fill=teal3, opacity=0.5, forget plot] 	fill between[of=upper and lower];
			\addplot[draw=teal1, thick, smooth] 			table[x index=0,y index=5] {fig/shkadov_1_env_no_bs.dat}; 
		
			\addplot [upper, forget plot] 				table[x index=0,y index=7] {fig/shkadov_2_envs_no_bs.dat}; 
			\addplot [lower, forget plot] 				table[x index=0,y index=6] {fig/shkadov_2_envs_no_bs.dat}; 
			\addplot [fill=red3, opacity=0.5, forget plot] 	fill between[of=upper and lower];
			\addplot[draw=red1, thick, smooth] 			table[x index=0,y index=5] {fig/shkadov_2_envs_no_bs.dat}; 
			
			\addplot [upper, forget plot] 				table[x index=0,y index=7] {fig/shkadov_4_envs_no_bs.dat}; 
			\addplot [lower, forget plot] 				table[x index=0,y index=6] {fig/shkadov_4_envs_no_bs.dat}; 
			\addplot [fill=gray3, opacity=0.5, forget plot] 	fill between[of=upper and lower];
			\addplot[draw=gray1, thick, smooth] 			table[x index=0,y index=5] {fig/shkadov_4_envs_no_bs.dat}; 
		
			\addplot [upper, forget plot] 				table[x index=0,y index=7] {fig/shkadov_8_envs_no_bs.dat}; 
			\addplot [lower, forget plot] 				table[x index=0,y index=6] {fig/shkadov_8_envs_no_bs.dat}; 
			\addplot [fill=blue3, opacity=0.5, forget plot] 	fill between[of=upper and lower];
			\addplot[draw=blue1, thick, smooth] 			table[x index=0,y index=5] {fig/shkadov_8_envs_no_bs.dat}; 
		
			\addplot [upper, forget plot] 					table[x index=0,y index=7] {fig/shkadov_16_envs_no_bs.dat}; 
			\addplot [lower, forget plot] 					table[x index=0,y index=6] {fig/shkadov_16_envs_no_bs.dat}; 
			\addplot [fill=bluegray3, opacity=0.5, forget plot] 	fill between[of=upper and lower];
			\addplot[draw=bluegray1, thick, smooth] 			table[x index=0,y index=5] {fig/shkadov_16_envs_no_bs.dat}; 
			
			\addplot [upper, forget plot] 					table[x index=0,y index=7] {fig/shkadov_32_envs_no_bs.dat}; 
			\addplot [lower, forget plot] 					table[x index=0,y index=6] {fig/shkadov_32_envs_no_bs.dat}; 
			\addplot [fill=orange3, opacity=0.5, forget plot] 		fill between[of=upper and lower];
			\addplot[draw=orange1, thick, smooth] 			table[x index=0,y index=5] {fig/shkadov_32_envs_no_bs.dat}; 
			
			\addplot [upper, forget plot] 					table[x index=0,y index=7] {fig/shkadov_64_envs_no_bs.dat}; 
			\addplot [lower, forget plot] 					table[x index=0,y index=6] {fig/shkadov_64_envs_no_bs.dat}; 
			\addplot [fill=purple3, opacity=0.5, forget plot] 		fill between[of=upper and lower];
			\addplot[draw=purple1, thick, smooth] 			table[x index=0,y index=5] {fig/shkadov_64_envs_no_bs.dat}; 
			
		\end{axis}
	\end{tikzpicture}
	\caption{Score against transitions, no bootstrapping}
	\label{fig:shkadov_parallel_transitions_no_bootstrap}
\end{subfigure}

\medskip

\begin{subfigure}[t]{.55\textwidth}
\centering
	\begin{tikzpicture}[	trim axis left, trim axis right, font=\scriptsize,
					upper/.style={	name path=upper, smooth, draw=none},
					lower/.style={	name path=lower, smooth, draw=none},
					limit/.style={	gray3, thick, dash pattern=on 2pt}]
		\begin{axis}[	xmin=0, xmax=90, ymin=-4, ymax=0, scale=0.8,
					width=\textwidth, height=5cm, scale only axis=true,
					every tick label/.append style={font=\scriptsize},
					legend cell align=left, legend pos=south east,
					grid=major, xlabel=training minutes, ylabel=score]

			\legend{1 env, 2 envs, 4 envs, 8 envs, 16 envs, 32 envs, 64 envs}

			\def\tun{5319}
			\addplot [upper, forget plot] 				table[x expr=\thisrowno{0}/\nsteps*\tun/60,y index=7] {fig/shkadov_1_env_semi_bs.dat}; 
			\addplot [lower, forget plot] 				table[x expr=\thisrowno{0}/\nsteps*\tun/60,y index=6] {fig/shkadov_1_env_semi_bs.dat}; 
			\addplot [fill=teal3, opacity=0.5, forget plot] 	fill between[of=upper and lower];
			\addplot[draw=teal1, thick, smooth] 			table[x expr=\thisrowno{0}/\nsteps*\tun/60,y index=5] {fig/shkadov_1_env_semi_bs.dat }; 
			\addplot[limit, forget plot]					coordinates {(\tun/60,-4) (\tun/60,0)};
		
			\def\ttwo{2834}
			\addplot [upper, forget plot] 				table[x expr=\thisrowno{0}/\nsteps*\ttwo/60,y index=7] {fig/shkadov_2_envs_semi_bs.dat}; 
			\addplot [lower, forget plot] 				table[x expr=\thisrowno{0}/\nsteps*\ttwo/60,y index=6] {fig/shkadov_2_envs_semi_bs.dat}; 
			\addplot [fill=red3, opacity=0.5, forget plot] 	fill between[of=upper and lower];
			\addplot[draw=red1, thick, smooth] 			table[x expr=\thisrowno{0}/\nsteps*\ttwo/60,y index=5] {fig/shkadov_2_envs_semi_bs.dat};
			\addplot[limit, forget plot]					coordinates {(\ttwo/60,-4) (\ttwo/60,0)};
			
			\def\tfour{1478}
			\addplot [upper, forget plot] 				table[x expr=\thisrowno{0}/\nsteps*\tfour/60,y index=7] {fig/shkadov_4_envs_semi_bs.dat}; 
			\addplot [lower, forget plot] 				table[x expr=\thisrowno{0}/\nsteps*\tfour/60,y index=6] {fig/shkadov_4_envs_semi_bs.dat}; 
			\addplot [fill=gray3, opacity=0.5, forget plot] 	fill between[of=upper and lower];
			\addplot[draw=gray1, thick, smooth] 			table[x expr=\thisrowno{0}/\nsteps*\tfour/60,y index=5] {fig/shkadov_4_envs_semi_bs.dat};
			\addplot[limit, forget plot]					coordinates {(\tfour/60,-4) (\tfour/60,0)};
		
			\def\thuit{777}
			\addplot [upper, forget plot] 				table[x expr=\thisrowno{0}/\nsteps*\thuit/60,y index=7] {fig/shkadov_8_envs_semi_bs.dat}; 
			\addplot [lower, forget plot] 				table[x expr=\thisrowno{0}/\nsteps*\thuit/60,y index=6] {fig/shkadov_8_envs_semi_bs.dat}; 
			\addplot [fill=blue3, opacity=0.5, forget plot] 	fill between[of=upper and lower];
			\addplot[draw=blue1, thick, smooth] 			table[x expr=\thisrowno{0}/\nsteps*\thuit/60,y index=5] {fig/shkadov_8_envs_semi_bs.dat}; 
			\addplot[limit, forget plot]					coordinates {(\thuit/60,-4) (\thuit/60,0)};
		
			\def\tseize{440}
			\addplot [upper, forget plot] 					table[x expr=\thisrowno{0}/\nsteps*\tseize/60,y index=7] {fig/shkadov_16_envs_semi_bs.dat}; 
			\addplot [lower, forget plot] 					table[x expr=\thisrowno{0}/\nsteps*\tseize/60,y index=6] {fig/shkadov_16_envs_semi_bs.dat}; 
			\addplot [fill=bluegray3, opacity=0.5, forget plot] 	fill between[of=upper and lower];
			\addplot[draw=bluegray1, thick, smooth] 			table[x expr=\thisrowno{0}/\nsteps*\tseize/60,y index=5] {fig/shkadov_16_envs_semi_bs.dat}; 
			\addplot[limit, forget plot]						coordinates {(\tseize/60,-4) (\tseize/60,0)};
			
			\def\ttdeux{306}
			\addplot [upper, forget plot] 				table[x expr=\thisrowno{0}/\nsteps*\ttdeux/60,y index=7] {fig/shkadov_32_envs_semi_bs.dat}; 
			\addplot [lower, forget plot] 				table[x expr=\thisrowno{0}/\nsteps*\ttdeux/60,y index=6] {fig/shkadov_32_envs_semi_bs.dat}; 
			\addplot [fill=orange3, opacity=0.5, forget plot] 	fill between[of=upper and lower];
			\addplot[draw=orange1, thick, smooth] 		table[x expr=\thisrowno{0}/\nsteps*\ttdeux/60,y index=5] {fig/shkadov_32_envs_semi_bs.dat}; 
			\addplot[limit, forget plot]					coordinates {(\ttdeux/60,-4) (\ttdeux/60,0)};
			
			\def\tsquatre{297}
			\addplot [upper, forget plot] 				table[x expr=\thisrowno{0}/\nsteps*\tsquatre/60,y index=7] {fig/shkadov_64_envs_semi_bs.dat}; 
			\addplot [lower, forget plot] 				table[x expr=\thisrowno{0}/\nsteps*\tsquatre/60,y index=6] {fig/shkadov_64_envs_semi_bs.dat}; 
			\addplot [fill=purple3, opacity=0.5, forget plot] 	fill between[of=upper and lower];
			\addplot[draw=purple1, thick, smooth] 		table[x expr=\thisrowno{0}/\nsteps*\tsquatre/60,y index=5] {fig/shkadov_64_envs_semi_bs.dat}; 
			\addplot[limit, forget plot]					coordinates {(\tsquatre/60,-4) (\tsquatre/60,0)};
			
		\end{axis}
	\end{tikzpicture}
	\caption{Score against walltime, EOE bootstrapping only}
	\label{fig:shkadov_parallel_walltime_semi_bootstrap}
\end{subfigure} \quad
\begin{subfigure}[t]{.35\textwidth}
\centering
	\begin{tikzpicture}[	trim axis left, trim axis right, font=\scriptsize,
					upper/.style={	name path=upper, smooth, draw=none},
					lower/.style={	name path=lower, smooth, draw=none},]
		\begin{axis}[	xmin=0, xmax=200000, ymin=-4, ymax=0, scale=0.8,
					xtick={0,50000,100000,150000,200000},
					xticklabels={$0$,$50k$,$100k$,$150k$,$200k$},
					scaled x ticks=false,
					width=\textwidth, height=5cm, scale only axis=true,
					every tick label/.append style={font=\scriptsize},
					legend cell align=left, legend pos=south east,
					grid=major, xlabel=transitions, ylabel={}]
				
			\addplot [upper, forget plot] 				table[x index=0,y index=7] {fig/shkadov_1_env_semi_bs.dat}; 
			\addplot [lower, forget plot] 				table[x index=0,y index=6] {fig/shkadov_1_env_semi_bs.dat}; 
			\addplot [fill=teal3, opacity=0.5, forget plot] 	fill between[of=upper and lower];
			\addplot[draw=teal1, thick, smooth] 			table[x index=0,y index=5] {fig/shkadov_1_env_semi_bs.dat}; 
		
			\addplot [upper, forget plot] 				table[x index=0,y index=7] {fig/shkadov_2_envs_semi_bs.dat}; 
			\addplot [lower, forget plot] 				table[x index=0,y index=6] {fig/shkadov_2_envs_semi_bs.dat}; 
			\addplot [fill=red3, opacity=0.5, forget plot] 	fill between[of=upper and lower];
			\addplot[draw=red1, thick, smooth] 			table[x index=0,y index=5] {fig/shkadov_2_envs_semi_bs.dat}; 
			
			\addplot [upper, forget plot] 				table[x index=0,y index=7] {fig/shkadov_4_envs_semi_bs.dat}; 
			\addplot [lower, forget plot] 				table[x index=0,y index=6] {fig/shkadov_4_envs_semi_bs.dat}; 
			\addplot [fill=gray3, opacity=0.5, forget plot] 	fill between[of=upper and lower];
			\addplot[draw=gray1, thick, smooth] 			table[x index=0,y index=5] {fig/shkadov_4_envs_semi_bs.dat}; 
		
			\addplot [upper, forget plot] 				table[x index=0,y index=7] {fig/shkadov_8_envs_semi_bs.dat}; 
			\addplot [lower, forget plot] 				table[x index=0,y index=6] {fig/shkadov_8_envs_semi_bs.dat}; 
			\addplot [fill=blue3, opacity=0.5, forget plot] 	fill between[of=upper and lower];
			\addplot[draw=blue1, thick, smooth] 			table[x index=0,y index=5] {fig/shkadov_8_envs_semi_bs.dat}; 
		
			\addplot [upper, forget plot] 					table[x index=0,y index=7] {fig/shkadov_16_envs_semi_bs.dat}; 
			\addplot [lower, forget plot] 					table[x index=0,y index=6] {fig/shkadov_16_envs_semi_bs.dat}; 
			\addplot [fill=bluegray3, opacity=0.5, forget plot] 	fill between[of=upper and lower];
			\addplot[draw=bluegray1, thick, smooth] 			table[x index=0,y index=5] {fig/shkadov_16_envs_semi_bs.dat}; 
			
			\addplot [upper, forget plot] 					table[x index=0,y index=7] {fig/shkadov_32_envs_semi_bs.dat}; 
			\addplot [lower, forget plot] 					table[x index=0,y index=6] {fig/shkadov_32_envs_semi_bs.dat}; 
			\addplot [fill=orange3, opacity=0.5, forget plot] 		fill between[of=upper and lower];
			\addplot[draw=orange1, thick, smooth] 			table[x index=0,y index=5] {fig/shkadov_32_envs_semi_bs.dat}; 
			
			\addplot [upper, forget plot] 					table[x index=0,y index=7] {fig/shkadov_64_envs_semi_bs.dat}; 
			\addplot [lower, forget plot] 					table[x index=0,y index=6] {fig/shkadov_64_envs_semi_bs.dat}; 
			\addplot [fill=purple3, opacity=0.5, forget plot] 		fill between[of=upper and lower];
			\addplot[draw=purple1, thick, smooth] 			table[x index=0,y index=5] {fig/shkadov_64_envs_semi_bs.dat}; 
			
		\end{axis}
	\end{tikzpicture}
	\caption{Score against transitions, EOE bootstrapping only}
	\label{fig:shkadov_parallel_transitions_semi_bootstrap}
\end{subfigure}

\medskip 

\begin{subfigure}[t]{.55\textwidth}
\centering
	\begin{tikzpicture}[	trim axis left, trim axis right, font=\scriptsize,
					upper/.style={	name path=upper, smooth, draw=none},
					lower/.style={	name path=lower, smooth, draw=none},
					limit/.style={	gray3, thick, dash pattern=on 2pt}]
		\begin{axis}[	xmin=0, xmax=90, ymin=-4, ymax=0, scale=0.8,
					width=\textwidth, height=5cm, scale only axis=true,
					every tick label/.append style={font=\scriptsize},
					legend cell align=left, legend pos=south east,
					grid=major, xlabel=training minutes, ylabel=score]
				
			\legend{1 env, 2 envs, 4 envs, 8 envs, 16 envs, 32 envs, 64 envs}

			\def\tun{5254}
			\addplot [upper, forget plot] 				table[x expr=\thisrowno{0}/\nsteps*\tun/60,y index=7] {fig/shkadov_1_env_bs.dat}; 
			\addplot [lower, forget plot] 				table[x expr=\thisrowno{0}/\nsteps*\tun/60,y index=6] {fig/shkadov_1_env_bs.dat}; 
			\addplot [fill=teal3, opacity=0.5, forget plot] 	fill between[of=upper and lower];
			\addplot[draw=teal1, thick, smooth] 			table[x expr=\thisrowno{0}/\nsteps*\tun/60,y index=5] {fig/shkadov_1_env_bs.dat}; 
			\addplot[limit, forget plot]					coordinates {(\tun/60,-4) (\tun/60,0)};
		
			\def\ttwo{2642}
			\addplot [upper, forget plot] 				table[x expr=\thisrowno{0}/\nsteps*\ttwo/60,y index=7] {fig/shkadov_2_envs_bs.dat}; 
			\addplot [lower, forget plot] 				table[x expr=\thisrowno{0}/\nsteps*\ttwo/60,y index=6] {fig/shkadov_2_envs_bs.dat}; 
			\addplot [fill=red3, opacity=0.5, forget plot] 	fill between[of=upper and lower];
			\addplot[draw=red1, thick, smooth] 			table[x expr=\thisrowno{0}/\nsteps*\ttwo/60,y index=5] {fig/shkadov_2_envs_bs.dat};
			\addplot[limit, forget plot]					coordinates {(\ttwo/60,-4) (\ttwo/60,0)};
			
			\def\tfour{1341}
			\addplot [upper, forget plot] 				table[x expr=\thisrowno{0}/\nsteps*\tfour/60,y index=7] {fig/shkadov_4_envs_bs.dat}; 
			\addplot [lower, forget plot] 				table[x expr=\thisrowno{0}/\nsteps*\tfour/60,y index=6] {fig/shkadov_4_envs_bs.dat}; 
			\addplot [fill=gray3, opacity=0.5, forget plot] 	fill between[of=upper and lower];
			\addplot[draw=gray1, thick, smooth] 			table[x expr=\thisrowno{0}/\nsteps*\tfour/60,y index=5] {fig/shkadov_4_envs_bs.dat};
			\addplot[limit, forget plot]					coordinates {(\tfour/60,-4) (\tfour/60,0)};
		
			\def\thuit{695}
			\addplot [upper, forget plot] 				table[x expr=\thisrowno{0}/\nsteps*\thuit/60,y index=7] {fig/shkadov_8_envs_bs.dat}; 
			\addplot [lower, forget plot] 				table[x expr=\thisrowno{0}/\nsteps*\thuit/60,y index=6] {fig/shkadov_8_envs_bs.dat}; 
			\addplot [fill=blue3, opacity=0.5, forget plot] 	fill between[of=upper and lower];
			\addplot[draw=blue1, thick, smooth] 			table[x expr=\thisrowno{0}/\nsteps*\thuit/60,y index=5] {fig/shkadov_8_envs_bs.dat}; 
			\addplot[limit, forget plot]					coordinates {(\thuit/60,-4) (\thuit/60,0)};
		
			\def\tseize{370}
			\addplot [upper, forget plot] 					table[x expr=\thisrowno{0}/\nsteps*\tseize/60,y index=7] {fig/shkadov_16_envs_bs.dat}; 
			\addplot [lower, forget plot] 					table[x expr=\thisrowno{0}/\nsteps*\tseize/60,y index=6] {fig/shkadov_16_envs_bs.dat}; 
			\addplot [fill=bluegray3, opacity=0.5, forget plot] 	fill between[of=upper and lower];
			\addplot[draw=bluegray1, thick, smooth] 			table[x expr=\thisrowno{0}/\nsteps*\tseize/60,y index=5] {fig/shkadov_16_envs_bs.dat}; 
			\addplot[limit, forget plot]						coordinates {(\tseize/60,-4) (\tseize/60,0)};
			
			\def\ttdeux{215}
			\addplot [upper, forget plot] 				table[x expr=\thisrowno{0}/\nsteps*\ttdeux/60,y index=7] {fig/shkadov_32_envs_bs.dat}; 
			\addplot [lower, forget plot] 				table[x expr=\thisrowno{0}/\nsteps*\ttdeux/60,y index=6] {fig/shkadov_32_envs_bs.dat}; 
			\addplot [fill=orange3, opacity=0.5, forget plot] 	fill between[of=upper and lower];
			\addplot[draw=orange1, thick, smooth] 		table[x expr=\thisrowno{0}/\nsteps*\ttdeux/60,y index=5] {fig/shkadov_32_envs_bs.dat}; 
			\addplot[limit, forget plot]					coordinates {(\ttdeux/60,-4) (\ttdeux/60,0)};
			
			\def\tsquatre{124}
			\addplot [upper, forget plot] 				table[x expr=\thisrowno{0}/\nsteps*\tsquatre/60,y index=7] {fig/shkadov_64_envs_bs.dat}; 
			\addplot [lower, forget plot] 				table[x expr=\thisrowno{0}/\nsteps*\tsquatre/60,y index=6] {fig/shkadov_64_envs_bs.dat}; 
			\addplot [fill=purple3, opacity=0.5, forget plot] 	fill between[of=upper and lower];
			\addplot[draw=purple1, thick, smooth] 		table[x expr=\thisrowno{0}/\nsteps*\tsquatre/60,y index=5] {fig/shkadov_64_envs_bs.dat}; 
			\addplot[limit, forget plot]					coordinates {(\tsquatre/60,-4) (\tsquatre/60,0)};
			
		\end{axis}
	\end{tikzpicture}
	\caption{Score against walltime, EOE + PT bootstrapping}
	\label{fig:shkadov_parallel_walltime_full_bootstrap}
\end{subfigure} \quad
\begin{subfigure}[t]{.35\textwidth}
\centering
	\begin{tikzpicture}[	trim axis left, trim axis right, font=\scriptsize,
					upper/.style={	name path=upper, smooth, draw=none},
					lower/.style={	name path=lower, smooth, draw=none},]
		\begin{axis}[	xmin=0, xmax=200000, ymin=-4, ymax=0, scale=0.8,
					xtick={0,50000,100000,150000,200000},
					xticklabels={$0$,$50k$,$100k$,$150k$,$200k$},
					scaled x ticks=false,
					width=\textwidth, height=5cm, scale only axis=true,
					every tick label/.append style={font=\scriptsize},
					legend cell align=left, legend pos=south east,
					grid=major, xlabel=transitions, ylabel={}]

			\addplot [upper, forget plot] 				table[x index=0,y index=7] {fig/shkadov_1_env_bs.dat}; 
			\addplot [lower, forget plot] 				table[x index=0,y index=6] {fig/shkadov_1_env_bs.dat}; 
			\addplot [fill=teal3, opacity=0.5, forget plot] 	fill between[of=upper and lower];
			\addplot[draw=teal1, thick, smooth] 			table[x index=0,y index=5] {fig/shkadov_1_env_bs.dat}; 
		
			\addplot [upper, forget plot] 				table[x index=0,y index=7] {fig/shkadov_2_envs_bs.dat}; 
			\addplot [lower, forget plot] 				table[x index=0,y index=6] {fig/shkadov_2_envs_bs.dat}; 
			\addplot [fill=red3, opacity=0.5, forget plot] 	fill between[of=upper and lower];
			\addplot[draw=red1, thick, smooth] 			table[x index=0,y index=5] {fig/shkadov_2_envs_bs.dat}; 
			
			\addplot [upper, forget plot] 				table[x index=0,y index=7] {fig/shkadov_4_envs_bs.dat}; 
			\addplot [lower, forget plot] 				table[x index=0,y index=6] {fig/shkadov_4_envs_bs.dat}; 
			\addplot [fill=gray3, opacity=0.5, forget plot] 	fill between[of=upper and lower];
			\addplot[draw=gray1, thick, smooth] 			table[x index=0,y index=5] {fig/shkadov_4_envs_bs.dat}; 
		
			\addplot [upper, forget plot] 				table[x index=0,y index=7] {fig/shkadov_8_envs_bs.dat}; 
			\addplot [lower, forget plot] 				table[x index=0,y index=6] {fig/shkadov_8_envs_bs.dat}; 
			\addplot [fill=blue3, opacity=0.5, forget plot] 	fill between[of=upper and lower];
			\addplot[draw=blue1, thick, smooth] 			table[x index=0,y index=5] {fig/shkadov_8_envs_bs.dat}; 
		
			\addplot [upper, forget plot] 					table[x index=0,y index=7] {fig/shkadov_16_envs_bs.dat}; 
			\addplot [lower, forget plot] 					table[x index=0,y index=6] {fig/shkadov_16_envs_bs.dat}; 
			\addplot [fill=bluegray3, opacity=0.5, forget plot] 	fill between[of=upper and lower];
			\addplot[draw=bluegray1, thick, smooth] 			table[x index=0,y index=5] {fig/shkadov_16_envs_bs.dat}; 
			
			\addplot [upper, forget plot] 					table[x index=0,y index=7] {fig/shkadov_32_envs_bs.dat}; 
			\addplot [lower, forget plot] 					table[x index=0,y index=6] {fig/shkadov_32_envs_bs.dat}; 
			\addplot [fill=orange3, opacity=0.5, forget plot] 		fill between[of=upper and lower];
			\addplot[draw=orange1, thick, smooth] 			table[x index=0,y index=5] {fig/shkadov_32_envs_bs.dat}; 
			
			\addplot [upper, forget plot] 					table[x index=0,y index=7] {fig/shkadov_64_envs_bs.dat}; 
			\addplot [lower, forget plot] 					table[x index=0,y index=6] {fig/shkadov_64_envs_bs.dat}; 
			\addplot [fill=purple3, opacity=0.5, forget plot] 		fill between[of=upper and lower];
			\addplot[draw=purple1, thick, smooth] 			table[x index=0,y index=5] {fig/shkadov_64_envs_bs.dat}; 
			
		\end{axis}
	\end{tikzpicture}
	\caption{Score against transitions, EOE + PT bootstrapping}
	\label{fig:shkadov_parallel_transitions_full_bootstrap}
\end{subfigure}
\caption{\textbf{Score curves obtained for different number of parallel environments.} (Top) With no bootstrapping (Middle) With end-of-episode (EOE) bootstrapping only (Bottom) With EOE and partial-trajectory (PT) bootstrapping}
\label{fig:shkadov_parallel}
\end{figure}

This section focuses on the interest of the partial-trajectory (PT) bootstrapping technique for parallel transitions collection proposed in section \ref{section:parallel}. To evaluate its interest, we consider the score curves obtained on the Shkadov environment in different configurations, the number of parallel environments ranging from $1$ to $64$ (see figure \ref{fig:shkadov_parallel}). First, we solve it with no bootstrapping at all (figures \ref{fig:shkadov_parallel_walltime_no_bootstrap} and \ref{fig:shkadov_parallel_transitions_no_bootstrap}), then with EOE bootstrapping only (figures \ref{fig:shkadov_parallel_walltime_semi_bootstrap} and \ref{fig:shkadov_parallel_transitions_semi_bootstrap}), then with EOE and PT bootstrapping (figures \ref{fig:shkadov_parallel_walltime_full_bootstrap} and \ref{fig:shkadov_parallel_transitions_full_bootstrap}). In all three cases, the score curves are plotted both against walltime and transitions, in order to visualize the parallel speedup obtained, as well as the relative performances using different number of parallel environments. For the clarity of the following discussion, we remind that $n_\text{env}$ designates the number of parallel environments used to gather transitions, and $n_\text{update}$ the number of full episodes to perform an update of the agent (here equal to 8). More, we introduce the notation $s^\text{M}_{n \rightarrow m}$ to designate the speedup observed for a parallel approach M when using $m$ parallel environments instead of $n$. Hence, for a perfect speedup, $s^\text{M}_{n \rightarrow m} = \frac{m}{n}$.

\begin{enumerate}
	\item \textbf{For $\bm{n_\text{env} \leq n_\text{update}}$}, the performance remains stable in each of the three configurations. This is expected, as the use of parallel environments only modifies the pace at which the updates occur, but the constitution of the update buffers is the same as it would have been for $n_\text{env} = 1$. Yet, the use of EOE bootstrapping leads to a faster convergence than the regular case, as was already observed in previous section. More, as PT bootstrapping does not occur when full episodes are used, figures \ref{fig:shkadov_parallel_transitions_semi_bootstrap} and \ref{fig:shkadov_parallel_transitions_full_bootstrap} present similar results for $n_\text{env} \leq 8$;
	\item \textbf{For $\bm{n_\text{env} > n_\text{update}}$}, a rapid decrease in convergence speed and final score is observed for the regular approach, with $n_\text{env} > 16$, resulting in very poor performance (figure \ref{fig:shkadov_parallel_transitions_no_bootstrap}). This illustrates the reasons that motivated the present contribution, \textit{i.e.} that the standard parallel paradigm results in impractical constraints that prevents massive environment parallelism. Introducing EOE bootstrapping (figure \ref{fig:shkadov_parallel_transitions_semi_bootstrap}) improves the situation by (i) generally speeding up the convergence, and (ii) improving the final performance of the agents, although the final score obtained for $n_\text{env} = 32$ remains sub-optimal, while that of $n_\text{env} = 64$ is poor. Additionally, the "flat steps" phenomenon described in \cite{rabaultkuhnle2019} clearly appears for $n_\text{env}=32$ and $64$, with the length of the steps being roughly equal to the number of transitions unrolled between two updates of the agent. When adding PT bootstrapping (figure \ref{fig:shkadov_parallel_transitions_full_bootstrap}), a clear improvement in the convergence speed is observed even for large $n_\text{env}$ values, and the gap in final performance significantly reduces. The flat steps phenomenon is still observed in the early stages of learning for $n_\text{env}=32$ and $64$, although with significantly reduced intensity.
	\item Similarly to \cite{rabaultkuhnle2019}, we observe decent speedups for $n_\text{env} \leq n_\text{update}$, with $s^\text{EOE+PT}_{1 \rightarrow 8} = 7.6$, against $s^\text{regular}_{1 \rightarrow 8} = 6.9$. This difference is attributed to the additional buffering overhead required in the regular approach, as more transitions are unrolled and stored between each update compared to PT bootstrapping. For higher number of parallel environments, we measure $s^\text{EOE+PT}_{1 \rightarrow 32} = 25.4$ against $s^\text{regular}_{1 \rightarrow 32} = 17.7$, and $s^\text{EOE+PT}_{1 \rightarrow 64} = 42.4$ against $s^\text{regular}_{1 \rightarrow 64} = 18.4$. Again, the excessive memory and buffering requirements of the regular case are the most probable cause of this discrepancy, as is evidenced by the fact that $s^\text{regular}_{1 \rightarrow 64} \simeq s^\text{regular}_{1 \rightarrow 32}$. In the present case, we underline that speedups could probably be improved by replacing the computation of a random initial state at each reset step of the environment by the loading of pre-computed initial states from files.
\end{enumerate}

Hence, the introduction of EOE and PT bootstrapping roughly allows to use 4 times more parallel environments than with the vanilla parallelism, thus reducing the resolution time of the Shkadov environment from nearly 1.5 hour to approximately 2 minutes while retaining the same final performance.

\section{Conclusion}

In the present contribution, we introduced a bootstrapped partial trajectory approach for parallel environments, in order to speed up learning for deep reinforcement learning agents while retaining their on-policiness. The proposed \blue{method} was tested on a CPU-intensive flow control case from the literature, bringing multiple improvements \blue{over regular approaches} such as (i) faster convergence, (ii) improved performance for $n_\text{env} > n_\text{update}$, (iii) improved parallel speedups, (iv) increased flexibility regarding the compatibility of $n_\text{env}$ and $n_\text{update}$, and (v) preserved on-policiness. The new parallel paradigm roughly allowed to safely exploit 4 times more parallel environment than the vanilla approach, and speedups as high as $42$ were measured. Such techniques, coupled with an efficient parallelism at the solver level, opens the door to the control of more complex, resource-demanding environments, thus pushing forward the current limits of DRL-based fluid flow control.

\section*{Acknowledgements}

The authors would like to thank A. Kuhnle for the fruitful discussions about bootstrapping.

\section*{Funding}

Funded/co-funded by the European Union (ERC, CURE, 101045042). Views and opinions expressed are however those of the author(s) only and do not necessarily reflect those of the European Union or the European Research Council. Neither the European Union nor the granting authority can be held responsible for them.

\appendix

\section{End-of-episode bootstrapping on \textsc{gym} environments}
\label{appendix:bootstrapping}

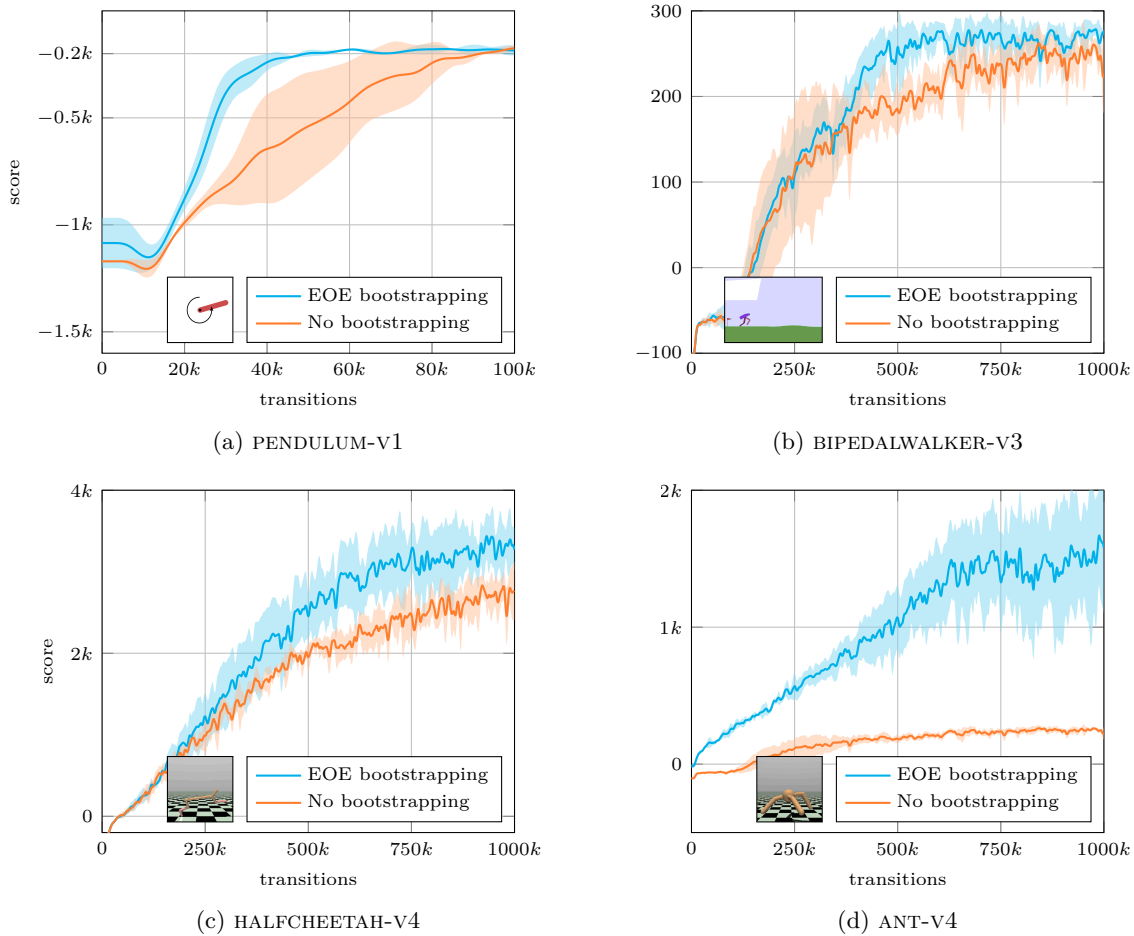
\begin{figure}
\centering
\begin{subfigure}[t]{.45\textwidth}
	\centering
	\begin{tikzpicture}[	trim axis left, trim axis right, font=\scriptsize,
					upper/.style={	name path=upper, smooth, draw=none},
					lower/.style={	name path=lower, smooth, draw=none},
					node distance=0.05cm]
		\begin{axis}[	xmin=0, xmax=100000, scale=0.8,
					ymin=-1600, ymax=0,
					scaled x ticks=false,
					xtick={0,20000,40000,60000,80000,100000},
					xticklabels={$0$,$20k$,$40k$,$60k$,$80k$,$100k$},
					ytick={-1500,-1000, -500,-200},
					yticklabels={$-1.5k$,$-1k$,$-0.5k$,$-0.2k$},
					legend cell align=left, legend pos=south east,
					legend style={name=leg},
					every tick label/.append style={font=\scriptsize},
					grid=major, xlabel=transitions, ylabel=score]
				
			\legend{EOE bootstrapping, No bootstrapping}

			\addplot [upper, forget plot] 				table[x index=0,y index=7] {fig/pendulum_bs.dat}; 
			\addplot [lower, forget plot] 				table[x index=0,y index=6] {fig/pendulum_bs.dat}; 
			\addplot [fill=blue3, opacity=0.5, forget plot] 	fill between[of=upper and lower];
			\addplot[draw=blue1, thick, smooth] 			table[x index=0,y index=5] {fig/pendulum_bs.dat}; 
		
			\addplot [upper, forget plot] 				table[x index=0,y index=7] {fig/pendulum_no_bs.dat}; 
			\addplot [lower, forget plot] 				table[x index=0,y index=6] {fig/pendulum_no_bs.dat}; 
			\addplot [fill=orange3, opacity=0.5, forget plot] 	fill between[of=upper and lower];
			\addplot[draw=orange1, thick, smooth] 		table[x index=0,y index=5] {fig/pendulum_no_bs.dat};
			
		\end{axis}
		{
		\setlength{\fboxrule}{0.4pt}%
		\node[left=of leg] (pic) {\fbox{\includegraphics[height=0.122\textwidth]{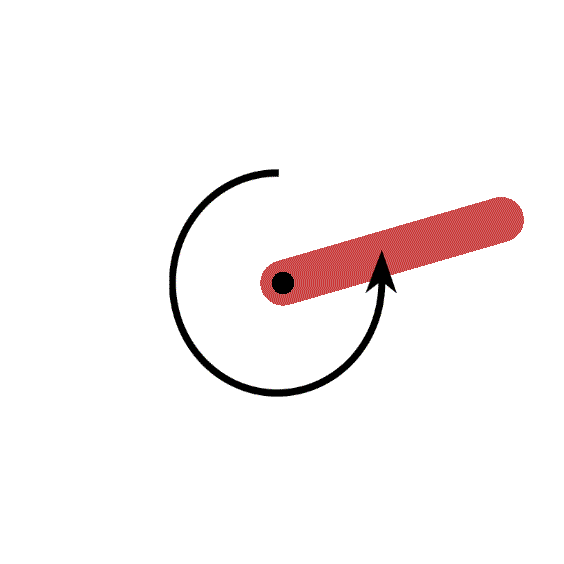}}};
		}
	\end{tikzpicture}
	\caption{\textsc{pendulum-v1}}
	\label{fig:pendulum_bootstrap_score}
\end{subfigure} \qquad
\begin{subfigure}[t]{.45\textwidth}
	\centering
	\begin{tikzpicture}[	trim axis left, trim axis right, font=\scriptsize,
					upper/.style={	name path=upper, smooth, draw=none},
					lower/.style={	name path=lower, smooth, draw=none},
					node distance=0.05cm]
		\begin{axis}[	xmin=0, xmax=1000000, scale=0.8,
					ymin=-100, ymax=300,
					scaled x ticks=false,
					xtick={0,250000,500000,750000,1000000},
					xticklabels={$0$,$250k$,$500k$,$750k$,$1000k$},
					ytick={-100,0,100,200,300},
					legend cell align=left, legend pos=south east,
					legend style={name=leg},
					every tick label/.append style={font=\scriptsize},
					grid=major, xlabel=transitions, ylabel={}]
				
			\legend{EOE bootstrapping, No bootstrapping}

			\addplot [upper, forget plot] 				table[x index=0,y index=7] {fig/bipedalwalker_bs.dat}; 
			\addplot [lower, forget plot] 				table[x index=0,y index=6] {fig/bipedalwalker_bs.dat}; 
			\addplot [fill=blue3, opacity=0.5, forget plot] 	fill between[of=upper and lower];
			\addplot[draw=blue1, thick, smooth] 			table[x index=0,y index=5] {fig/bipedalwalker_bs.dat}; 
		
			\addplot [upper, forget plot] 				table[x index=0,y index=7] {fig/bipedalwalker_no_bs.dat}; 
			\addplot [lower, forget plot] 				table[x index=0,y index=6] {fig/bipedalwalker_no_bs.dat}; 
			\addplot [fill=orange3, opacity=0.5, forget plot] 	fill between[of=upper and lower];
			\addplot[draw=orange1, thick, smooth] 		table[x index=0,y index=5] {fig/bipedalwalker_no_bs.dat};
			
		\end{axis}
		{
		\setlength{\fboxrule}{0.4pt}%
		\node[left=of leg] (pic) {\fbox{\includegraphics[height=0.122\textwidth]{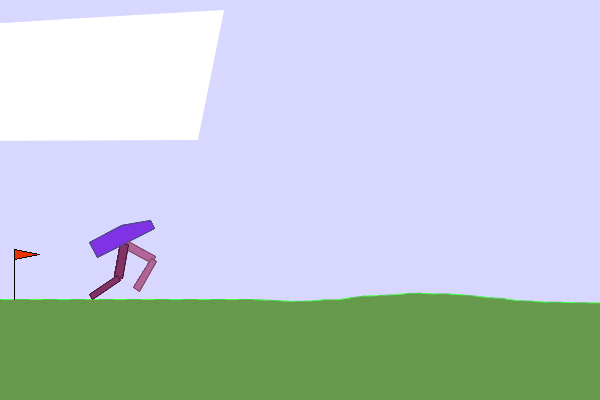}}};
		}
	\end{tikzpicture}
	\caption{\textsc{bipedalwalker-v3}}
	\label{fig:bipedalwalker_bootstrap_score}
\end{subfigure}

\medskip

\begin{subfigure}[t]{.45\textwidth}
	\centering
	\begin{tikzpicture}[	trim axis left, trim axis right, font=\scriptsize,
					upper/.style={	name path=upper, smooth, draw=none},
					lower/.style={	name path=lower, smooth, draw=none},
					node distance=0.05cm]
		\begin{axis}[	xmin=0, xmax=1000000, scale=0.8,
					ymin=-200, ymax=4000,
					scaled x ticks=false,
					xtick={0,250000,500000,750000,1000000},
					xticklabels={$0$,$250k$,$500k$,$750k$,$1000k$},
					ytick={0,2000,4000},
					yticklabels={$0$,$2k$,$4k$},
					legend cell align=left, legend pos=south east,
					legend style={name=leg},
					every tick label/.append style={font=\scriptsize},
					grid=major, xlabel=transitions, ylabel=score]
				
			\legend{EOE bootstrapping, No bootstrapping}

			\addplot [upper, forget plot] 				table[x index=0,y index=7] {fig/halfcheetah_bs.dat}; 
			\addplot [lower, forget plot] 				table[x index=0,y index=6] {fig/halfcheetah_bs.dat}; 
			\addplot [fill=blue3, opacity=0.5, forget plot] 	fill between[of=upper and lower];
			\addplot[draw=blue1, thick, smooth] 			table[x index=0,y index=5] {fig/halfcheetah_bs.dat}; 
		
			\addplot [upper, forget plot] 				table[x index=0,y index=7] {fig/halfcheetah_no_bs.dat}; 
			\addplot [lower, forget plot] 				table[x index=0,y index=6] {fig/halfcheetah_no_bs.dat}; 
			\addplot [fill=orange3, opacity=0.5, forget plot] 	fill between[of=upper and lower];
			\addplot[draw=orange1, thick, smooth] 		table[x index=0,y index=5] {fig/halfcheetah_no_bs.dat};
			
		\end{axis}
		{
		\setlength{\fboxrule}{0.4pt}%
		\node[left=of leg] (pic) {\fbox{\includegraphics[height=0.122\textwidth]{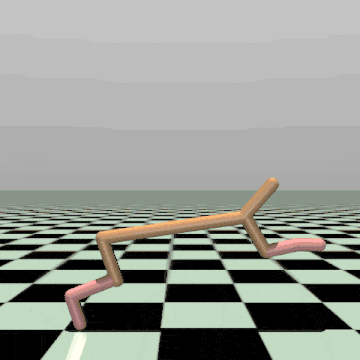}}};
		}
	\end{tikzpicture}
	\caption{\textsc{halfcheetah-v4}}
	\label{fig:halfcheetah_bootstrap_score}
\end{subfigure} \qquad
\begin{subfigure}[t]{.45\textwidth}
	\centering
	\begin{tikzpicture}[	trim axis left, trim axis right, font=\scriptsize,
					upper/.style={	name path=upper, smooth, draw=none},
					lower/.style={	name path=lower, smooth, draw=none},
					node distance=0.05cm]
		\begin{axis}[	xmin=0, xmax=1000000, scale=0.8,
					ymin=-500, ymax=2000,
					scaled x ticks=false,
					xtick={0,250000,500000,750000,1000000},
					xticklabels={$0$,$250k$,$500k$,$750k$,$1000k$},
					ytick={0,1000,2000},
					yticklabels={$0$,$1k$,$2k$},
					legend cell align=left, legend pos=south east,
					legend style={name=leg},
					every tick label/.append style={font=\scriptsize},
					grid=major, xlabel=transitions, ylabel={}]
				
			\legend{EOE bootstrapping, No bootstrapping}

			\addplot [upper, forget plot] 				table[x index=0,y index=7] {fig/ant_bs.dat}; 
			\addplot [lower, forget plot] 				table[x index=0,y index=6] {fig/ant_bs.dat}; 
			\addplot [fill=blue3, opacity=0.5, forget plot] 	fill between[of=upper and lower];
			\addplot[draw=blue1, thick, smooth] 			table[x index=0,y index=5] {fig/ant_bs.dat}; 
		
			\addplot [upper, forget plot] 				table[x index=0,y index=7] {fig/bipedalwalker_no_bs.dat}; 
			\addplot [lower, forget plot] 				table[x index=0,y index=6] {fig/bipedalwalker_no_bs.dat}; 
			\addplot [fill=orange3, opacity=0.5, forget plot] 	fill between[of=upper and lower];
			\addplot[draw=orange1, thick, smooth] 		table[x index=0,y index=5] {fig/bipedalwalker_no_bs.dat};
			
		\end{axis}
		{
		\setlength{\fboxrule}{0.4pt}%
		\node[left=of leg] (pic) {\fbox{\includegraphics[height=0.122\textwidth]{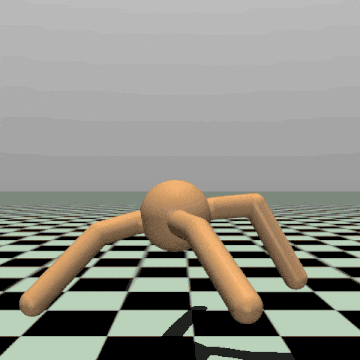}}};
		}
	\end{tikzpicture}
	\caption{\textsc{ant-v4}}
	\label{fig:ant_bootstrap_score}
\end{subfigure}
\caption{\textbf{Illustration of the interest of end-of-episode bootstrapping on continuous control problems} from the \textsc{gym} and \textsc{mujoco} libraries, solved with the PPO algorithm. (Top left) The \textsc{pendulum-v1} environment, in which the agent controls the torque applied to the pendulum junction, the goal being to keep the pole balanced vertically. The score comparison is made over 500 episodes (equivalent to \num{100000} transitions). (Top right) The \textsc{bipedalwalker-v3} environment, in which the agent controls the 4 torques applied at the hips and knees of the walker. The score comparison is made over \num{1000000} transitions. (Bottom left) The \textsc{halfcheetah-v4} environment, where the agent learns to run with a cat-like robot, using 6 torques. The score comparison is made over \num{1000000} transitions. (Bottom right) The \textsc{ant-v4} environment, where a four-leg ant learns to move using 8 torques. The score comparison is made over \num{1000000} transitions. For all environments, the solid color line indicates the average over the 5 runs, while the light-colored area around it indicates the standard deviation around the average.}
\label{fig:bootstrap_score}
\end{figure}

This appendix illustrates the interest of bootstrapping on continuous control environments from the \textsc{gym} package, such as \textsc{pendulum-v1} and \textsc{bipedalwalker-v3}, and on locomotion problems from the \textsc{mujoco} package, such as \textsc{halfcheetah-v4} and \textsc{ant-v4}. For the sake of brevity, the environments are not fully described here, and the interested reader is referred to the original publications for details \cite{gym, mujoco}. On figure \ref{fig:bootstrap_score}, the scores obtained with the PPO method are compared with and without the bootstrapping technique, averaged over 5 runs. As can be observed, bootstrapping the end-of-episode target in the case of time-outs leads to a largely improved convergence speed, as well as in a lower variability between different runs.

\newpage

\section{\blue{Convergence of the numerical discretization}}
\label{appendix:convergence}

\blue{The numerical discretization proposed to solve the system (\ref{eq:shkadov}) is tested by using a manufactured solution based on the following \textit{a priori} expressions for $h(x,t)$ and $q(x,t)$:}

\begin{equation}
\label{eq:manufactured}
\begin{split}
	h(x,t) 	&= \cos\left( \omega t - k x \right), \\
	q(x,t)		&= \frac{\omega}{k} \sin\left( \omega t - k x \right),
\end{split}
\end{equation}

\begin{figure}[h!]
\centering
\def\nsteps{200000}
\begin{tikzpicture}[	trim axis left, trim axis right, font=\scriptsize,
				upper/.style={	name path=upper, smooth, draw=none},
				lower/.style={	name path=lower, smooth, draw=none},
				node distance=0.05cm]
	\begin{loglogaxis}[	xmin=0.01, xmax=1.0, ymin=0.03, ymax=200, scale=0.8,
					every tick label/.append style={font=\scriptsize},
					legend cell align=left, legend pos=south east,
					grid=both, xlabel=$\Delta x$, ylabel={error}]
				
		\addplot[draw=blue1, thick, smooth, mark=*] coordinates {(0.5, 124.3) (0.2, 19.3) (0.1, 4.6) (0.05, 1.14) (0.025, 0.28) (0.0125, 0.070)};
		\draw[black] (0.1,4.6) -- (0.2, 4.6) -- (0.2, 19.3);
		\node[] at (0.23, 9.0) {2};
		\node[] at (0.145, 3.2) {1};

	\end{loglogaxis}
\end{tikzpicture}
\caption{\textbf{Convergence in space for the discretization of the Shkadov system} obtained with a manufactured solution.}
\label{fig:manufactured}
\end{figure}
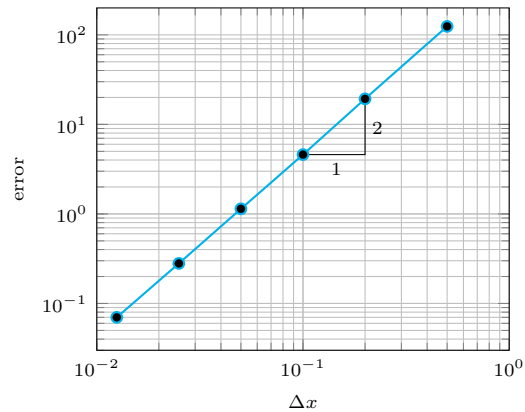

\blue{with $\omega = 2.3$ and $k=0.77$. The discretization error induced by the derivatives computation is measured and plotted as a function of the spatial discretization step $\Delta x$ in figure \ref{fig:manufactured}. As can be observed, a second order convergence in obtained, which correlates with the chosen numerical scheme in the absence of sharp gradients.}

\section{Solved Shkadov environment}
\label{appendix:solved_env}

\begin{figure}[h!]
\centering
\pgfdeclarelayer{background}
\pgfsetlayers{background,main}

\begin{subfigure}[t]{\textwidth}
	\centering
	\begin{tikzpicture}[	scale=0.7, trim axis left, trim axis right, font=\scriptsize]
		\begin{axis}[	xmin=0, xmax=220, ymin=0, ymax=2, scale=1.0,
					xtick={0,50,100,150,200},
					width=\textwidth, height=.15\textwidth, scale only axis=true,
					legend cell align=left, legend pos=north east,
					grid=major, ylabel=$h$]
				
		\def\x{150}
		\def\w{1.5}
		\def\s{10}

		\draw[fill=green2, draw=gray2] 			(\x+0*\s-\w,1) rectangle (\x+0*\s+\w,1+0.49);
		\draw[fill=green2, draw=gray2] 			(\x+1*\s-\w,1) rectangle (\x+1*\s+\w,1-0.81);
		\draw[fill=green2, draw=gray2] 			(\x+2*\s-\w,1) rectangle (\x+2*\s+\w,1-0.83);
		\draw[fill=green2, draw=gray2] 			(\x+3*\s-\w,1) rectangle (\x+3*\s+\w,1-0.76);
		\draw[fill=green2, draw=gray2] 			(\x+4*\s-\w,1) rectangle (\x+4*\s+\w,1-0.23);
		
		\addplot[draw=gray1, very thick, smooth] 	table[x index=0,y index=1] {fig/field_0.dat};
			
		\end{axis}
	\end{tikzpicture}
    	\caption{$t=0$, start of control}
	\label{fig:shkadov_fields_0}
\end{subfigure}
\begin{subfigure}[t]{\textwidth}
	\centering
	\begin{tikzpicture}[	scale=0.7, trim axis left, trim axis right, font=\scriptsize]
		\begin{axis}[	xmin=0, xmax=220, ymin=0, ymax=2, scale=1.0,
					xtick={0,50,100,150,200},
					width=\textwidth, height=.15\textwidth, scale only axis=true,
					legend cell align=left, legend pos=north east,
					grid=major, ylabel=$h$]
				
			\def\x{150}
			\def\w{1.5}
			\def\s{10}
		
			\draw[fill=green2, draw=gray2] 			(\x+0*\s-\w,1) rectangle (\x+0*\s+\w,1+0.029);
			\draw[fill=green2, draw=gray2] 			(\x+1*\s-\w,1) rectangle (\x+1*\s+\w,1-0.050);
			\draw[fill=green2, draw=gray2] 			(\x+2*\s-\w,1) rectangle (\x+2*\s+\w,1+0.18);
			\draw[fill=green2, draw=gray2] 			(\x+3*\s-\w,1) rectangle (\x+3*\s+\w,1-0.24);
			\draw[fill=green2, draw=gray2] 			(\x+4*\s-\w,1) rectangle (\x+4*\s+\w,1+0.054);
		
			\addplot[draw=gray1, very thick, smooth] 	table[x index=0,y index=1] {fig/field_100.dat};
			
		\end{axis}
	\end{tikzpicture}
    	\caption{$t=100$}
	\label{fig:shkadov_fields_100}
\end{subfigure}
\begin{subfigure}[t]{\textwidth}
	\centering
	\begin{tikzpicture}[	scale=0.7, trim axis left, trim axis right, font=\scriptsize]
		\begin{axis}[	xmin=0, xmax=220, ymin=0, ymax=2, scale=1.0,
					xtick={0,50,100,150,200},
					width=\textwidth, height=.15\textwidth, scale only axis=true,
					legend cell align=left, legend pos=north east,
					grid=major, ylabel=$h$]
				
			\def\x{150}
			\def\w{1.5}
			\def\s{10}
		
			\draw[fill=green2, draw=gray2] 			(\x+0*\s-\w,1) rectangle (\x+0*\s+\w,1-0.033);
			\draw[fill=green2, draw=gray2] 			(\x+1*\s-\w,1) rectangle (\x+1*\s+\w,1-0.035);
			\draw[fill=green2, draw=gray2] 			(\x+2*\s-\w,1) rectangle (\x+2*\s+\w,1+0.049);
			\draw[fill=green2, draw=gray2] 			(\x+3*\s-\w,1) rectangle (\x+3*\s+\w,1-0.027);
			\draw[fill=green2, draw=gray2] 			(\x+4*\s-\w,1) rectangle (\x+4*\s+\w,1+0.0007);
		
			\addplot[draw=gray1, very thick, smooth] 	table[x index=0,y index=1] {fig/field_200.dat};
			
		\end{axis}
	\end{tikzpicture}
    	\caption{$t=200$}
	\label{fig:shkadov_fields_200}
\end{subfigure}
\begin{subfigure}[t]{\textwidth}
	\centering
	\begin{tikzpicture}[	scale=0.7, trim axis left, trim axis right, font=\scriptsize]
		\begin{axis}[	xmin=0, xmax=220, ymin=0, ymax=2, scale=1.0,
					xtick={0,50,100,150,200},
					width=\textwidth, height=.15\textwidth, scale only axis=true,
					legend cell align=left, legend pos=north east,
					grid=major, ylabel=$h$]
				
			\def\x{150}
			\def\w{1.5}
			\def\s{10}
		
			\draw[fill=green2, draw=gray2] 			(\x+0*\s-\w,1) rectangle (\x+0*\s+\w,1+0.087);
			\draw[fill=green2, draw=gray2] 			(\x+1*\s-\w,1) rectangle (\x+1*\s+\w,1-0.065);
			\draw[fill=green2, draw=gray2] 			(\x+2*\s-\w,1) rectangle (\x+2*\s+\w,1+0.066);
			\draw[fill=green2, draw=gray2] 			(\x+3*\s-\w,1) rectangle (\x+3*\s+\w,1+0.028);
			\draw[fill=green2, draw=gray2] 			(\x+4*\s-\w,1) rectangle (\x+4*\s+\w,1+0.011);
		
			\addplot[draw=gray1, very thick, smooth] 	table[x index=0,y index=1] {fig/field_300.dat};
			
		\end{axis}
	\end{tikzpicture}
    	\caption{$t=300$}
	\label{fig:shkadov_fields_300}
\end{subfigure}
\begin{subfigure}[t]{\textwidth}
	\centering
	\begin{tikzpicture}[	scale=0.7, trim axis left, trim axis right, font=\scriptsize]
		\begin{axis}[	xmin=0, xmax=220, ymin=0, ymax=2, scale=1.0,
					xtick={0,50,100,150,200},
					width=\textwidth, height=.15\textwidth, scale only axis=true,
					legend cell align=left, legend pos=north east,
					grid=major, ylabel=$h$]
				
			\def\x{150}
			\def\w{1.5}
			\def\s{10}
		
			\draw[fill=green2, draw=gray2] 			(\x+0*\s-\w,1) rectangle (\x+0*\s+\w,1-0.059);
			\draw[fill=green2, draw=gray2] 			(\x+1*\s-\w,1) rectangle (\x+1*\s+\w,1-0.056);
			\draw[fill=green2, draw=gray2] 			(\x+2*\s-\w,1) rectangle (\x+2*\s+\w,1+0.081);
			\draw[fill=green2, draw=gray2] 			(\x+3*\s-\w,1) rectangle (\x+3*\s+\w,1+0.067);
			\draw[fill=green2, draw=gray2] 			(\x+4*\s-\w,1) rectangle (\x+4*\s+\w,1-0.001);
		
			\addplot[draw=gray1, very thick, smooth] 	table[x index=0,y index=1] {fig/field_400.dat};
			
		\end{axis}
	\end{tikzpicture}
    	\caption{$t=400$}
	\label{fig:shkadov_fields_400}
\end{subfigure}
\caption{\textbf{Evolution of the flow under control of the agent, using 5 jets.} The jets strengths are represented with green rectangles.}
\label{fig:shkadov_fields}
\end{figure}
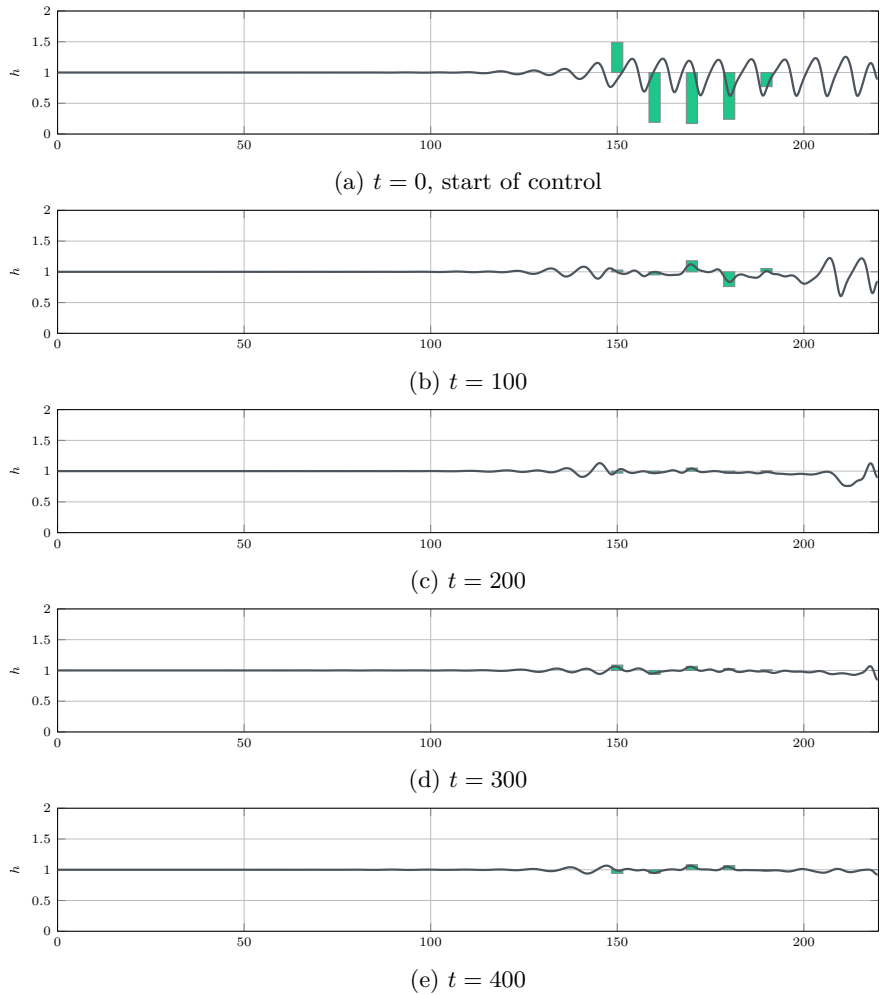

In figure \ref{fig:shkadov_fields}, we present the evolution of the field in time under the control of a solved agent for 5 jets using the default parameters. As can be observed, the agent quickly constrains the height of the fluid around $h=1$, before entering a quasi-stationary state in which a set of minimal, quasi constant jet actuations keeps the flow from developing instabilities.

\clearpage

\bibliographystyle{unsrt}
\bibliography{bib}

\end{document}

%% file: packages.tex
\usepackage[utf8]{inputenc}
\usepackage{arxiv}
\usepackage{float}
\usepackage{enumitem}
\usepackage[english]{babel}
\usepackage{amssymb}
\usepackage{amsmath,mathtools}
\usepackage[table]{xcolor}
\usepackage{graphicx}
\usepackage{algorithmicx}
\usepackage{algpseudocode}
\usepackage{algorithm}
\usepackage{multirow}
\usepackage{overpic}
\usepackage{psfrag}
\usepackage{bm}
\usepackage{lineno}
\usepackage{url}
\usepackage{lipsum}
\usepackage{epsfig}
\usepackage{marvosym}
\usepackage{verbatim}
\usepackage{subcaption}
\usepackage{caption}
\usepackage{hyperref}
\hypersetup{colorlinks=true,linkcolor=red,citecolor=blue}
\usepackage{ulem}
\usepackage{lmodern}
\usepackage{booktabs} 
\usepackage{moresize} 
\usepackage{siunitx}
\usepackage{mathtools}
\usepackage{tikz}
\usepackage{pgfplots}
\pgfplotsset{compat=1.14}
\usepgfplotslibrary{fillbetween}
\usetikzlibrary{shapes, arrows, arrows.meta, calc, positioning, decorations,
decorations.pathmorphing, decorations.text, spy}
\usepackage{tikz-3dplot}
\pgfplotsset{translate gnuplot=true}
\usepackage{etex} 								
\usepgfplotslibrary{external} 						
\tikzsetexternalprefix{ext/}							
\usepackage{booktabs}
\usepackage{makecell}
\usepackage{rotating,tabularx}

%% file: environment.tex
\definecolor{blue1}		{RGB}{0,177,234}				
\definecolor{blue2}		{RGB}{76,200,239}				
\definecolor{blue3}		{RGB}{127,215,244}				
\definecolor{blue4}		{RGB}{178,231,248}				
\definecolor{bluegray1}	{RGB}{0,127,167}				
\definecolor{bluegray2}	{RGB}{76,165,193}				
\definecolor{bluegray3}	{RGB}{127,191,211}				
\definecolor{bluegray4}	{RGB}{178,216,228}				
\definecolor{gray1}		{RGB}{76,84,93}				
\definecolor{gray2}		{RGB}{129,135,141}				
\definecolor{gray3}		{RGB}{165,169,174}				
\definecolor{gray4}		{RGB}{201,203,206}				
\definecolor{gray5}		{RGB}{230,230,230}				
\definecolor{gray6}		{RGB}{245,245,245}				
\definecolor{orange1}	{RGB}{255,126,46}				
\definecolor{orange2}	{RGB}{255,164,108}				
\definecolor{orange3}	{RGB}{255,190,150}				
\definecolor{orange4}	{RGB}{255,216,192}				
\definecolor{brown1}		{RGB}{205,133,63}				
\definecolor{green1}		{RGB}{0,168,107}				
\definecolor{green2}		{RGB}{30,198,137}				
\definecolor{green3}		{RGB}{60,228,167}				
\definecolor{green4}		{RGB}{90,235,167}				
\definecolor{purple1}		{RGB}{89,89,171}				
\definecolor{purple2}		{RGB}{129,129,195}				
\definecolor{purple3}		{RGB}{189,189,225}				
\definecolor{purple4}		{RGB}{209,209,255}				
\definecolor{teal1}		{RGB}{0,128,128}				
\definecolor{teal2}		{RGB}{80,158,158}				
\definecolor{teal3}		{RGB}{130,198,198}				
\definecolor{teal4}		{RGB}{160,228,228}				
\definecolor{red1}		{RGB}{255,0,0}					
\definecolor{red2}		{RGB}{255,30,30}				
\definecolor{red3}		{RGB}{255,60,60}				
\definecolor{red4}		{RGB}{255,90,90}				
\definecolor{royal1}		{RGB}{2,119,189}				
\definecolor{royal2}		{RGB}{32,149,219}				
\definecolor{royal3}		{RGB}{62,179,249}				
\definecolor{royal4}		{RGB}{92,209,255}				

\newcommand\blue[1]{\textcolor{black}{#1}}

\pgfplotsset{
    colormap={custom_map}{[5pt]
            rgb255(0pt)=(255,126,46);
            rgb255(500pt)=(255,190,150);
            rgb255(1000pt)=(0,177,234);
            rgb255(1500pt)=(127,215,244);
    },
}

\renewcommand{\emph}[1]{\textit{#1}}
\newcommand{\eps}{\varepsilon}								
\setenumerate{label=\textcolor{bluegray1}{$\bm{\diamond}$},itemsep=0.0pt,topsep=5pt}

\renewcommand{\Re}{\operatorname{\mathit{R\kern-.04em e}}} 
\newcommand{\Nu}{\operatorname{\mathit{N\kern-.15em u}}} 
\newcommand{\Ra}{\operatorname{\mathit{R\kern-.04em a}}} 

\algnewcommand{\IIf}[3]{\State #1 \algorithmicif\ #2\ \algorithmicelse\ #3}

\tikzset{
	mStyle/.style={row sep=-2*\pgflinewidth, column sep=-2*\pgflinewidth},
	mWH/.style n args={2}{minimum width=#1cm, minimum height=#2cm},
	mNode/.style n args={1}{draw=blue1, thick, fill=#1, mWH},
	mRounded/.style n args={5}{append after command={\pgfextra
        		\draw[mNode={#5}, sharp corners]%
    		(\tikzlastnode.west)%
    		[rounded corners=#1pt] |- (\tikzlastnode.north)%
    		[rounded corners=#2pt] -| (\tikzlastnode.east)%
    		[rounded corners=#3pt] |- (\tikzlastnode.south)%
    		[rounded corners=#4pt] -| (\tikzlastnode.west);%
   		\endpgfextra}},
	mTL/.style n args={1}{mRounded={3}{0}{0}{0}{#1}},
	mTR/.style n args={1}{mRounded={0}{3}{0}{0}{#1}},
	mBL/.style n args={1}{mRounded={0}{0}{0}{3}{#1}},
	mBR/.style n args={1}{mRounded={0}{0}{3}{0}{#1}},
	mL/.style n args={1}{mRounded={3}{0}{0}{3}{#1}},
	mR/.style n args={1}{mRounded={0}{3}{3}{0}{#1}},
	mN/.style n args={1}{mRounded={0}{0}{0}{0}{#1}},
	mB/.style n args={1}{mRounded={0}{0}{3}{3}{#1}},
	mT/.style n args={1}{mRounded={3}{3}{0}{0}{#1}}
}

\tikzset{arrow/.style={thick,gray2,-stealth}}

\tikzset{label/.style={text depth=.3\baselineskip, text height=.5\baselineskip}}